%% file: paper_neurips_2024.tex
\documentclass{article}

\PassOptionsToPackage{numbers, compress}{natbib}

\usepackage[final]{neurips_2024}

\usepackage[utf8]{inputenc} 
\usepackage[T1]{fontenc}    
\usepackage{hyperref}       
\usepackage{url}            
\usepackage{booktabs}       
\usepackage{amsfonts}       
\usepackage{nicefrac}       
\usepackage{microtype}      
\usepackage{xcolor}         
\usepackage{csquotes}       
\newcommand{\todo}[1]{{}}
\usepackage{algorithm}
\usepackage{algpseudocode}
\usepackage{amsmath}
\usepackage{amssymb}
\usepackage{graphicx}
\usepackage{wrapfig}
\usepackage{subcaption}
\usepackage{multirow}
\usepackage[most]{tcolorbox}

\input{lib}

\input{math}

\title{Generating Highly Designable Proteins with \\
Geometric Algebra Flow Matching}

\author{%
Simon Wagner$^{*\,1,2}$
\quad
Leif Seute$^{*\,1,2,3}$\thanks{Equal contribution}
\quad
Vsevolod Viliuga$^{1,3,4}$
\quad
Nicolas Wolf$^{1,2,3}$\\
\textbf{Frauke Gräter}$^{1,2,3}$
\quad
\textbf{Jan Stühmer}$^{1,5}$ \\
\\$^1$Heidelberg Institute for Theoretical Studies, Heidelberg, Germany\\
$^2$IWR, Heidelberg University, Heidelberg, Germany\\
$^3$Max Planck Institute for Polymer Research, Mainz, Germany\\
$^4$SciLifeLab and DBB at Stockholm University, Stockholm, Sweden\\
$^5$IAR, Karlsruhe Institute of Technology, Karlsruhe, Germany
}

\begin{document}

\maketitle

\input{0_abstract}

\input{1_introduction}
\input{2_background}

\input{3_method}

\input{4_experiments}
\input{5_conclusion}

\newpage
\pagebreak
\bibliographystyle{plainnat} 
\bibliography{bibliography_neurips_2024}

\newpage
\pagebreak


\appendix

\setcounter{figure}{0}
\renewcommand\thefigure{\thesection.\arabic{figure}} 
\input{6_appendices}


\newpage
\pagebreak
\input{7_checklist}


\end{document}

%% file: lib.tex
\newcommand{\eqlabel}[1]{\label{eq:#1}}
\renewcommand{\eqref}[1]{\ref{eq:#1}}

\newcommand{\figlabel}[1]{\label{fig:#1}}
\newcommand{\figref}[1]{\ref{fig:#1}}

\newcommand{\alglabel}[1]{\label{alg:#1}}

\newcommand{\seclabel}[1]{\label{sec:#1}}
\newcommand{\secref}[1]{\ref{sec:#1}}

\usepackage{xspace}
\makeatletter
\DeclareRobustCommand\onedot{\futurelet\@let@token\@onedot}
\def\@onedot{\ifx\@let@token.\else.\null\fi\xspace}

\def\ie{\emph{i.e}\onedot}

\makeatother

%% file: math.tex
\newcommand{\myvector}[1]{\vec{\mathbf{#1}}}

\newcommand{\gavec}[1]{\mathbf{#1}}
\newcommand{\multivector}[1]{\mathbf{#1}}

\newcommand{\pga}{\mathbb{G}_{3,0,1}}

\newcommand{\deflabel}[1]{\label{def:#1}}
\newcommand{\thmlabel}[1]{\label{thm:#1}}

\newcommand{\defref}[1]{Definition~\ref{def:#1}}

\newcommand{\defemph}[1]{\textbf{#1}}

\usepackage{xparse}

\NewDocumentCommand{\cliff}{g g g}{
    \IfNoValueTF{#1}{
        \mathbb{G}
    }{
        \IfNoValueTF{#2}{
            \mathbb{G}_{#1}
        }{
            \IfNoValueTF{#3}{
                \mathbb{G}_{{#1},{#2}}
            }{
                \mathbb{G}_{{#1},{#2},{#3}}
            }
        }
    }
}

\newcommand{\mvec}[1]{\mathbf{#1}}
\newcommand{\evec}[1]{\vec{\mathbf{#1}}}

\newcommand{\reverse}[1]{\widetilde{#1}}

\newcommand{\egadual}[1]{{#1}^{\ast}}

\newcommand{\egaI}{\mvec{I}}

\newcommand{\pgaI}{\mathcal{I}}

\newcommand{\grade}[2]{\left\langle #1 \right\rangle_{#2}}

\usepackage{dsfont}

\DeclareMathOperator{\spn}{span}

\newif\ifusetcolorbox
\usetcolorboxtrue 

\newtcolorbox{dbox}[2][attach title to upper={\ }]
{colback=blue!10!white,
coltitle=black,
fonttitle=\bfseries,
enhanced,
every float=\centering,
frame hidden,
title=#2,#1}

\newcounter{theorem} 
\renewcommand{\thetheorem}{\thesection.\arabic{theorem}}

\NewDocumentEnvironment{theorem}{m +b}{
    \refstepcounter{theorem}
    \ifusetcolorbox
        \begin{dbox}{Theorem \thetheorem: #1}
            #2
        \end{dbox}
    \else
        \textbf{#1:} #2
    \fi
}{}

\newcounter{definition} 
\renewcommand{\thedefinition}{\thesection.\arabic{definition}}

\NewDocumentEnvironment{mathdef}{m +b}{
    \refstepcounter{definition}
    \ifusetcolorbox
        \begin{dbox}{Definition \thedefinition: #1}
            #2
        \end{dbox}
    \else
        \textbf{#1:} #2
    \fi
}{}

\definecolor{dblue}{RGB}{100, 100, 204}

%% file: 0_abstract.tex
\begin{abstract}
We introduce a generative model for protein backbone design utilizing geometric products and higher order message passing.
In particular, we propose Clifford Frame Attention (CFA), an extension of the invariant point attention (IPA) architecture from AlphaFold2, in which the backbone
residue frames and geometric features are represented in the projective geometric algebra.
This enables to construct geometrically expressive messages between residues, including higher order terms, using the bilinear operations of the algebra.
We evaluate our architecture by incorporating it into the framework of FrameFlow, a state-of-the-art flow matching model for protein backbone generation.
The proposed model achieves high designability, diversity and novelty, while also sampling protein backbones that follow the statistical distribution of secondary structure elements found in naturally occurring proteins,
a property so far only insufficiently achieved by many state-of-the-art generative models.
\end{abstract}

%% file: 1_introduction.tex
\section{Introduction}


%
Recent years have shown tremendous progress in applying deep learning to computational chemistry, where applications of learning-based approaches have enabled unprecedented progress across a broad range of problems, such as molecular property prediction~\citep{thomas2018tensor,gasteiger2019directional,satorras2021gnn,brandstetter2022geometric}, protein-ligand docking~\citep{corso2023diffdock}, protein structure prediction~\citep{jumper2021highly, baek2023efficient, lin2023evolutionary}, and \textit{de novo} protein design~\citep{watson2023novo,yim2023framediff,bose2023se,yim2023frameflow,lin2023genie}.
In case of protein design, state-of-the-art methods typically represent the structure of a protein of $N$ residues as an element of \text{SE}(3)$^{N}$, \ie as a collection of $N$ frames, each of which describes the position and orientation of an individual protein residue.
Among the most successful methods are those based on diffusion models~\cite{sohl2015deep,song2020score,watson2023novo,yim2023framediff,lin2023genie,anand2022protein} and flow matching~\citep{lipman2022flow,chen2023riemannian,tong2024improving,yim2023frameflow}, which make use of architectures that incorporate the invariant point attention (IPA) of AlphaFold2~\citep{jumper2021highly}.
By modeling a protein through the frames of its backbone, the task of protein structure generation is reformulated as to model the distribution of a set of frames, an inherently \emph{geometric} problem.
%

Advances in geometric deep learning have lead to architectures that are equivariant towards rotations and translations~\cite{bronstein2017geometric,cohen2021equivariant,bronstein2021geometric}, which can be regarded as a geometric inductive bias that enhances performance and data efficiency.
Most generative models for protein design achieve equivariance by using geometric features that are expressed in the canonical local coordinate frames representing the protein backbone.
The coordinates of these features are thus invariant with respect to global rotations and translations, which allows to apply general layers and non-linearities.

Another widely used approach to construct equivariant architectures is to embed internal features in symmetry group representations and restrict neural network operations to equivariant functions \citep{thomas2018tensor,gasteiger2019directional,satorras2021gnn,brandstetter2022geometric,batzner2022,batatia2022mace,liao2023equiformer,simeon2023tensornet}.
Explicit equivariance in these models has usually been limited to the orthogonal group $O(3)$, which contains rotations and reflections, while most models are instead invariant towards translations. Only recently,~\citet{brehmer2023geometric} proposed an architecture based on the framework of \citet{ruhe2023clifford} that enables explicit equivariance towards both translations and rotations by utilizing the projective geometric algebra (PGA)~\citep{gunn2020projective, dorst2020guided}.

\begin{figure}[t]
    \centering
    \includegraphics[width=1.0\linewidth]{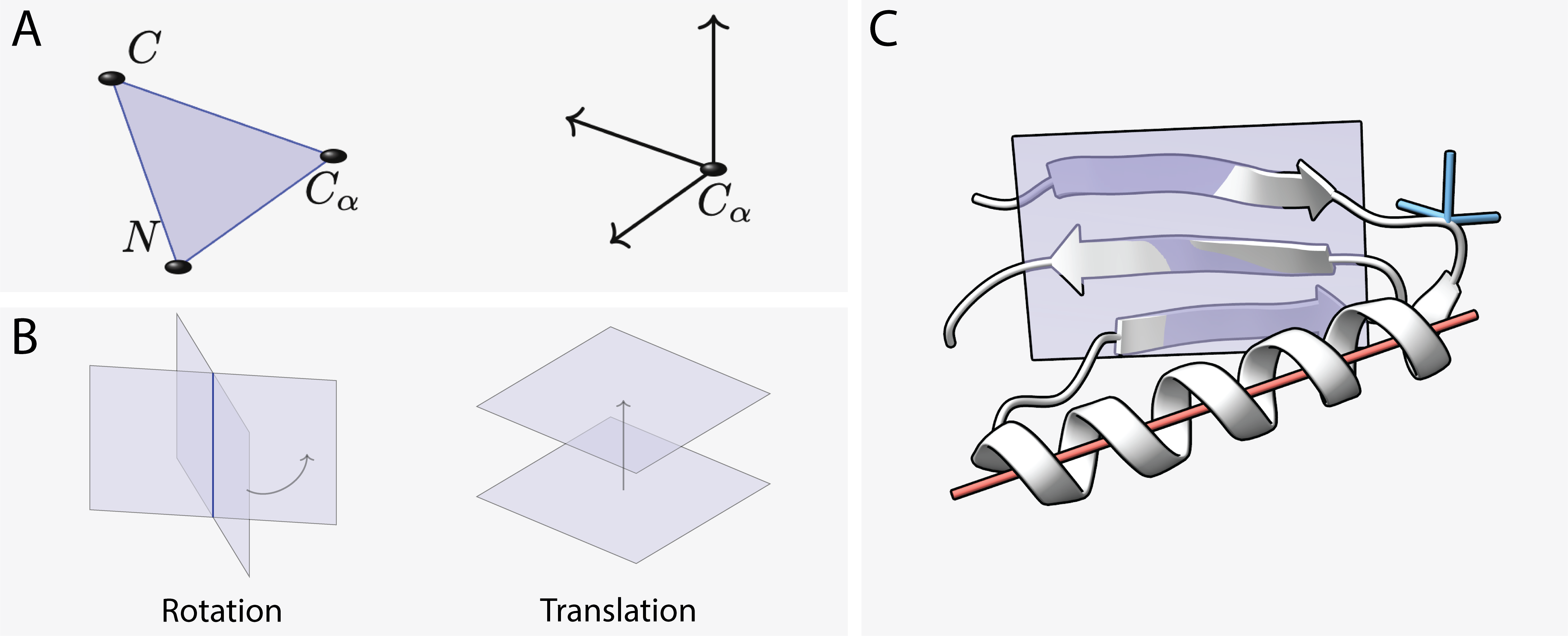}
    \caption{(\textbf{A}) Protein backbone residue with three backbone atoms represented by a coordinate frame. (\textbf{B}) In PGA, a frame can be represented via the geometric product of four planes. Two of the planes parameterize the frame's rotation around their line of intersection, while the other two encode the frame's translation along the separation vector between them.
    (\textbf{C}) An exemplary protein backbone structure containing an $\alpha$-helix and a $\beta$-sheet. Lines (red), planes (violet) and Euclidean frames (blue) can all be embedded as elements of PGA, facilitating a geometric inductive bias for learning representations of the abstract geometry of the protein.\vspace{-0.32cm}}
    \label{fig:pga_frames}
\end{figure}

Inspired by their work, we demonstrate that apart from its use as $\text{E}(3)$-equivariant formalism, PGA provides a powerful framework for representing the geometry of protein backbones when incorporated into the local frame formulation of protein design architectures.
While we utilize PGA to explicitly represent the frames of the protein backbone as elements of the algebra,
PGA additionally provides a strong inductive bias as its elements can represent abstract geometric objects like points, lines and planes, which are well suited to capture the geometry of secondary structure elements, such as $\alpha$-helices and $\beta$-sheets (Figure~\ref{fig:pga_frames}).
Moreover, bilinear operations of the algebra enable to compute many geometric relations between those objects, such as distances, angles, projections and incidences.
\paragraph{Main contributions:}
We introduce \emph{Clifford frame attention} (CFA), an extension of the invariant point attention (IPA) architecture of AlphaFold2~\citep{jumper2021highly} by expressing geometric features as elements of the projective geometric algebra and using its bilinear operations to construct geometrically expressive, higher order messages.
We incorporate CFA into an existing flow matching framework for protein backbone generation, FrameFlow~\citep{yim2023frameflow}, and demonstrate in our experiments that the proposed method achieves state-of-the-art performance in the combination of designability, diversity and novelty of generated protein samples.
While, especially for small proteins, other models with high designability often over-represent $\alpha$-helices, the proposed method captures the broad distribution of secondary structures of naturally occurring proteins, which we believe is crucial for designing proteins with vast functionalities.
Notably, since IPA is widely used throughout the field, CFA may be readily included in many other protein-related machine learning models.

\subsection{Related Work}

\paragraph{Geometric (Clifford) algebra in neural networks}
Neural networks that use the Clifford algebra were first proposed by~\citet{pearson1994neural}, an extension of the multi layer perceptron (MLP) by Clifford algebras, which was later studied further by~\citet{buchholz2008clifford}.
More recently,~\citet{ruhe2023geometric} propose Geometric Clifford Algebra Networks, using geometric (Clifford) algebras based on their Clifford Neural Layers~\citep{brandstetter2023clifford}, and extend this framework in~\citep{ruhe2023clifford} to $\text{E}(3)$-equivariant representations.
\citet{brehmer2023geometric} propose to use projective geometric algebra, which enables $\text{SE}(3)$-equivariant feature representations, and the representation of frames as elements of the algebra.

\paragraph{Generative models for protein design}
\citet{watson2023novo} propose RFdifffusion, a generative model for protein backbone design that utilizes the pre-trained protein structure prediction network RoseTTAFold~\citep{baek2021accurate}.
\citet{yim2023framediff} propose FrameDiff, which defines a diffusion model over a set of frames, $\text{SE}(3)^N$, and extend this method within the flow matching framework~\citep{yim2023frameflow}.
\citet{bose2023se} propose FoldFlow and its variants, diffusion and flow matching models over $\text{SE}(3)^N$.
\citet{lin2023genie} propose an equivariant encoder and decoder architecture.
\citet{wu2024protein} train a transformer model to predict angles between adjacent residues.
\citet{mao2023modeling} propose vector field networks, which is also an extension of IPA, and employ them in FrameDiff. The main difference to our approach is that VFN uses virtual atoms as geometric features and \emph{vector field operators} to model interactions, whereas CFA use multivectors and the geometric bilinears of PGA respectively.

%% file: 2_background.tex
\section{Background}
\seclabel{background}

This section provides an introduction to the mathematical frameworks that we refer to in this paper, flow matching and Geometric algebra, and a brief introduction to protein design.

\subsection{Geometric Algebra}
\seclabel{geometric-algebra}

A Clifford algebra over a real vector space, typically referred to as geometric algebra, is a powerful mathematical framework for describing geometric objects including points, lines, planes and operations on these objects in an algebraically concise way~\citep{hestenes2012clifford,dorst2007geometric}.

In more technical terms, given a vector space $V$ and a quadratic form $q: V \xrightarrow{} \mathbb{F}$ from the vector space to the underlying field $\mathbb{F}$, we can construct a geometric algebra as the unitary, associative, non-commutative algebra with the property $\gavec{v}^2 = q(\gavec{v})$ for every $\gavec{v} \in V$~\citep{ruhe2023geometric}. $\gavec{v}^2 = \gavec{v}\gavec{v}$ denotes the \emph{geometric product}, the bilinear operation of the algebra, of the vector $\gavec{v}$ with itself. In the geometric context, $q$ may be thought of as the metric of the vector space, meaning that for a vector $\gavec{v}$, $q(\gavec{v})$ is its squared norm. Elements of the algebra are called \emph{multivectors}. They can be constructed by forming geometric products between basis vectors of $V$ and linearly combining them.

In this paper we are mainly interested in the projective geometric algebra (PGA)~\citep{gunn2020projective,dorst2020guided}, which we denote as $\pga$. It is the geometric algebra over $V = \mathbb{R}^4$ with vector basis 
\begin{equation}
    \gavec{e_0}, \gavec{e_1}, \gavec{e_2}, \gavec{e_3} \; \text{and} \; q(\gavec{e_1}) = q(\gavec{e_2}) =  q(\gavec{e_3}) = 1, \; q(\gavec{e_0}) = 0 \, .
\end{equation}
The full algebra has 16 basis elements, which can be grouped in grades according to the number of vector basis elements they are constructed from. Following~\citet{gunn2020projective}, we can interpret elements of different grades as geometric objects such as planes, lines and points in $3D$ space, as listed in Table~\ref{tab:pga_elements} and visualized in Figure \ref{fig:pga_primitives}. Working in a four dimensional space has the advantage that we are able to represent lines and planes that do not necessarily include the origin, which is crucial for the description of translations as shown below. 

Central to the proposed method is the fact that PGA allows the representation of elements of the Euclidean group $\mathrm{E}(3)$ as elements of the algebra. Given a plane $\gavec{p} \in \pga$, an arbitrary geometric object $\multivector{X} \in \pga$ can be reflected across the plane via the sandwich product,
\begin{equation}
    \multivector{X}' = \gavec{p} \multivector{\hat{X}} \gavec{p} \, ,
    \label{eq:pga_reflection_main}
\end{equation}
where $\multivector{\hat{X}}$ is the grade involution that flips the sign of elements with odd grade. A vector in PGA may thus be interpreted both as a plane and as a reflection operator. One can extend this idea using the Cartan-Dieudonné theorem, which states that any $\mathrm{E}(3)$ transformation can be represented by repeated reflections. Two consecutive reflections through intersecting planes result in a rotation around the line of intersection and two reflections through two parallel planes correspond to a translation along the separation vector of the planes. The associated elements of PGA are obtained by taking the geometric product of the respective planes. We thus represent each residue frame of a protein by a multivector, a so called \emph{motor}, corresponding to a rotation followed by a translation as shown in Figure~\ref{fig:pga_frames}. Similar to eq.~\ref{eq:pga_reflection_main}, a motor $\multivector{M} \in \pga$ can be applied to an arbitrary element $\multivector{X} \in \pga$ according to
\begin{equation}
    \multivector{X}' = \multivector{M} \multivector{X} \multivector{M}^{-1} \, .
\end{equation}
We provide a more detailed introduction to geometric algebra in Appendix \ref{sec:ga_background}

\subsection{Flow Matching}
Flow matching~\citep{lipman2022flow}, as a generalization of diffusion models~\citep{sohl2015deep,song2020score}, offers a framework for learning continuous normalizing flows (CNFs)~\cite{chen2018neural}, $\phi\colon[0,1]\times\Omega\to\Omega$ that transform a general prior distribution $p_0$, defined on the domain $\Omega$, to a target distribution $p_1$ by evaluating the probability path
\begin{equation}
    p_t=[\phi_t]_*p_0
    \label{eq:prob-path}
\end{equation}
at $t=1$, where $*$ denotes the pushforward.
The flow $\phi_t$ can be expressed in terms of a vector field $v_t$ by solving the ordinary differential equation (ODE)
\begin{equation}
    \label{eqn:CNF_ODE}
    \frac{\mathrm{d}}{\mathrm{d}t} \phi_t(x) = v_t(\phi_t(x)), \quad\phi_0(x)=x \, .
\end{equation}
If the vector field $v_t$ is known, one can generate samples from $p_1$ by sampling from the prior and integrating the flow ODE.
Lipman et al.~\cite{lipman2022flow} showed that $v_t$ can be learned by regressing on a vector field $u_t(x|x_1)$ that is conditioned on a data sample $x_1\sim p_1$, i.e. by minimizing the loss function
\begin{equation}
    \label{eq:flow-mat-loss}
    \mathcal{L} = \mathbb{E}_{t\sim\mathcal{U}(0,1),x_1\sim p_1,x_t\sim p_t(\cdot|x_1)}\left[\left\|v_t(x_t)-u_t(x_t|x_1)\right\|^2\right] \, .
\end{equation}
Here, $p_t(x|x_1)$ is the conditional probability path induced by $u_t(x_t|x_1)$, which is commonly chosen as linear interpolation between the prior sample $x_0$ and the target $x_1$ by setting the conditional flow to
\begin{equation}
    \psi_t(x_0|x_1) \equiv (1-t) x_0 + t x_1 \, ,
\end{equation}
from which $u_t(x|x_1)$ can be obtained by forming the time derivative.
As shown by~\cite{lipman2022flow,chen2023riemannian}, sampling from $p_t(x_t|x_1)$ in eq.~\ref{eq:flow-mat-loss} can then be realized by transforming a sample $x_0$ from the prior $p_0$ to $x_t\equiv\psi_t(x_0|x_1)$.
This formulation in terms of conditional distributions allows to learn dynamic optimal transport (OT) plans from prior to target using minibatch OT as shown by~\citet{tong2024improving}.

The framework of CNFs and flow matching can be generalized for sampling from distributions on Riemannian manifolds $\mathcal{M}$, such as $\text{SE}(3)^N$, as described by~\citet{chen2023riemannian}.
Then, the flow $\phi\colon[0,1]\times\mathcal{M}\to\mathcal{M}$ is a diffeomorphism, and a vector field is learned as a smooth map $u_t\colon[0,1]\times\mathcal{M}\to\mathcal{TM}$ to the tangent space $\mathcal{TM}$.
A natural choice for the map $\psi_t(x_0|x_1)$ on a Riemannian manifold is the connecting path with minimal length, the geodesic.

\subsection{Flow Matching for Protein Structures}
\label{sec:flow-matching-general}

For sampling from the distribution of protein backbones, we represent each protein residue by a frame $T=(r,x)\in\mathrm{SE}(3)$ that corresponds to a translation $x\in\mathbb{R}^3$ and rotation $r\in\mathrm{SO}(3)$, thus the domain for the flow matching process is $\mathcal{M}\equiv\text{SE}(3)^N$, as described in FrameDiff~\cite{yim2023framediff}.
The frame for a given residue is defined through the backbone atoms $[N,C_\alpha,C]$, where the translation $x$ is chosen as the displacement of the $C_\alpha$ atom relative to the origin, while the rotation $r$ is constructed using a Gram-Schmidt process~\cite{jumper2021highly} on the vectors $[C-C_\alpha,N-C_\alpha]$.

Similar to~\citep{yim2023frameflow}, we define the conditional flow $\psi_t(T_0|T_1)$ as the geodesic between $T_0$ and $T_1$,
\begin{equation}
    \psi_t(T_0|T_1)=\exp_{T_0}(t\log_{T_0}(T_1))\,.
\end{equation}
We calculate the time derivative of the flow as described in FrameFlow~\cite{yim2023frameflow} and regress on it as in eq.~\ref{eq:flow-mat-loss}.
As prior $p_0$, we choose a $3N$-dimensional normal distribution with unit variance $\mathcal{N}(0,I)$ for translations and, for rotations, adopt the heuristic trick of using $\mathcal{IG}SO(3)$ during training and $\mathcal{U}(SO(3))$ for inference from~\citet{yim2023frameflow}.
Additionally, we use minibatch optimal transport \cite{tong2024improving} with the equivariant cost function from \cite{klein2023equivariant, bose2023se}.

\subsection{Designability of Proteins and Importance of their Structural Composition} \label{sec:bg_designability}

Until recently, computational tools for protein design have relied on physics-based energy functions~\citep{schymkowitz2005foldx, leaver2011rosetta3, webb2016comparative}, restricting protein design campaigns to pre-defined topologies, geometries, and secondary structure elements.
In contrast, deep generative models 
learn the probability distribution of protein structures from the training set, enabling less constrained sampling of the structure space.

Ideally, the main goal of any generative model should be the creation of structurally diverse and novel proteins, thereby maximizing the access to unseen structures and possibly functions. Since the functionality of proteins is grounded in their structure, it is crucial to extensively cover the broad range of \emph{secondary structure} elements and topologies found in natural proteins.
Typically, the properties of generated proteins are assessed by the scores of
\emph{diversity} and \emph{novelty}, indicating how similar generated backbones are to each other, and to known native proteins, respectively.
While explicit analyses of secondary structure elements of generated proteins are often overlooked,~\citep{lin2023genie, wu2024protein} suggest that diffusion models commonly generate redundant protein structures composed mostly of $\alpha$-helices, which raises the question of how much functionality can be hypothetically encoded in these proteins.

A key quality check for generated proteins is their biophysical consistency in the sense that their sequence, indeed, folds into the intended structure~\cite{ovchinnikov2021structure}.
In the case of backbone generation, this is typically evaluated by predicting a sequence from the generated structure with the inverse folding model ProteinMPNN~\citep{dauparas2022protmpnn}.
The obtained sequence is re-folded with ESMFold~\citep{lin2023evolutionary} and if the resulting structure aligns well with the original backbone, the latter is called \emph{designable}~\cite{trippe2022diffusion, watson2023novo}.
Thus, the designability metric measures consistency between well-established sequence and structure prediction models, which can be seen as a proxy for biophysical validity and practical realizability \cite{roney2022state}.

%% file: 3_method.tex
\section{Geometric Algebra Flow Matching for Protein Backbone Generation}
\seclabel{methods}

We propose a geometric algebra-based neural network architecture to predict the vector fields in a flow matching process on $\text{SE}(3)^N$ and call this approach Geometric Algebra Flow Matching (GAFL).
The proposed neural network architecture is an extension of the one introduced in FrameDiff~\cite{yim2023framediff}, where we replace its central component, the invariant point attention (IPA) block from AlphaFold2~\cite{jumper2021highly} by \emph{Clifford Frame Attention} (CFA), which we explain in more detail in the next paragraph.

Input features are the noised frames $T_t$, pairwise spatial distances, positional encodings of absolute and relative sequence positions, and the flow matching time $t$. As FrameDiff, we use self-conditioning.
The network relies on a series of six blocks, in which the frames are updated consecutively to predict the denoised backbone structure and from this the respective conditional vector fields.
Each block uses CFA to perform message passing between protein residues, in particular processing geometric information as given by geometric node features and the current set of frames. The $\text{SE}(3)$ invariant output of CFA is fed through an MLP and a transformer~\cite{vaswani2017attention} and then used to predict frame updates.
For a detailed explanation of the architecture, we refer to Appendix \ref{sec:framediff_architecture}.

\subsection*{Clifford Frame Attention}
\label{gafl-model}

The original IPA mechanism uses geometric node features in the form of $3\text{D}$ points for the calculation of attention scores and as queries, keys and values, as described in Appendix~\ref{sec:originalipa}.
The features are expressed in local coordinate frames, which allows the use of arbitrary layers without breaking equivariance.
Messages between nodes are constructed as a linear combination of attention values, weighted by attention scores.
While other generative models for protein design incorporate the original version of IPA directly~\cite{yim2023framediff, yim2023frameflow, bose2023se}, we propose to enhance its geometric expressivity by performing the following modifications.

\paragraph{PGA features:}

We replace the point-valued attention values by multivectors, which can encode points, lines, planes and Euclidean frames as shown in Figure \ref{fig:pga_frames}. At the same time we decrease the number of channels to retain approximately the same number of parameters.

\paragraph{Geometric messages:} 

Instead of linearly combining geometric features in the message passing step, we construct messages utilizing the geometric bilinear layer that was introduced as part of the Geometric Algebra Transformer in~\citet{brehmer2023geometric}, effectively calculating geometric product and join, another bilinear operation of PGA (see \defref{join}), between the node features $\mvec{V}_i$ and $\mvec{V}_j$ (see Algorithm \ref{alg:pairwisegeometricbilinear}). These two operations are able to compute many geometric relations as detailed in Appendix \ref{sec:ga_background} and \cite{dorst2020guided}. The messages $m_{ij}^{h,p}$ from node $i$ to node $j$ can then be written as

\begin{equation}
    m_{ij}^{h,p} = \operatorname{GeometricBilinear}\left(\mvec{T}_i^{-1}\left(\mvec{T}_j\mvec{V}_j^{h,p}\mvec{T}_j^{-1}\right)\mvec{T}_i, \, \mvec{V}_i^{h,p}\right),
\end{equation}

where $\mvec{T}_i, \mvec{T}_j$ are the frame transformations of the corresponding node, which ensure that the bilinear operations are performed in the same reference frame. When aggregating the messages we make use of the bilinearity of the products to exchange the sum and the bilinear layer to obtain:
\begin{equation}
\sum_j a_{ij}^h m_{ij}^{h,p} = \operatorname{GeometricBilinear}\Big(\mvec{T}_i^{-1}\sum_j a_{ij}^h \left(\mvec{T}_j\mvec{V}_j^{h,p}\mvec{T}_j^{-1}\right)\mvec{T}_i, \, \mvec{V}_i^{h,p}\big) \, ,\eqlabel{bilinear_contraction}
\end{equation}
The operation is thus performed on a node level and scales linearly with the number of nodes.

\paragraph{Higher order message passing:} We also incorporate higher order messages into the proposed architecture, that is, messages that depend on more than two nodes and are thus capable of describing relationships beyond pairwise interaction.
As in~\citep{batatia2022mace}, we use bilinearity of geometric product and join to construct higher order messages by multiplying aggregated two-body messages $m$ and $m^\prime$, e.g.
\begin{equation}
    \label{eq:higher_order}
    \left( \sum_j m^{}_{ij}\right) \left(\sum_k m^\prime_{ik}\right)
    =
    \sum_{jk}m^{}_{ij}m^\prime_{ik}
    \equiv
    \sum_{jk}m^{(3)}_{ijk}
\end{equation}
can be seen as sum of three-body messages $m^{(3)}_{ijk}$.
In practice, we use a learnable linear combination of versions of eq.~\ref{eq:higher_order} projected onto different multivector grades using the geometric product and the join, as described in algorithm~\ref{alg:geometricmanybodycontraction}.

\paragraph{Frames as features:} Since we embed both residue frames and geometric node features in $\cliff{3}{0}{1}$, it is a natural idea to combine them on a node level such that they can interact via the geometric bilinears during message passing. To this end we compute relative frame transformations for all pairs and aggregate them with the attention weights,
\begin{equation}
    \multivector{T}^\text{rel}_i \equiv \sum_{j}a_{ij} \multivector{T}^{-1}_i\multivector{T}_j\,.
    \label{eq:T-rel-main}
\end{equation}
We concatenate these with the remaining geometric features before the construction of geometric messages and also pass them directly to the backbone update block using a residual connection.
The full CFA algorithm is provided in Appendix~\ref{sec:cfa-method}.

Although PGA would in principle allow to construct a fully equivariant architecture~(\cite{brehmer2023geometric}) without using local frames, we decide to keep the local frame formulation of IPA, since it allows to use more general layers and non-linearities.
Moreover, the common problem of ambiguous local frame choices that other architectures suffer from \cite{wang2022graph} is not apparent in our case since backbone residues provide a canonical, geometrically meaningful choice for the local frames.
We discuss the equivariance of GAFL in Appendix \ref{sec:gafl_equivariance}.

%% file: 4_experiments.tex
\newcommand{\error}[3]{{#1}^{\, #2}_{\, #3}} 

\section{Experiments}
\seclabel{experiments}

\begin{figure}[h]
    \centering
    \includegraphics[page=1,width=1\textwidth]{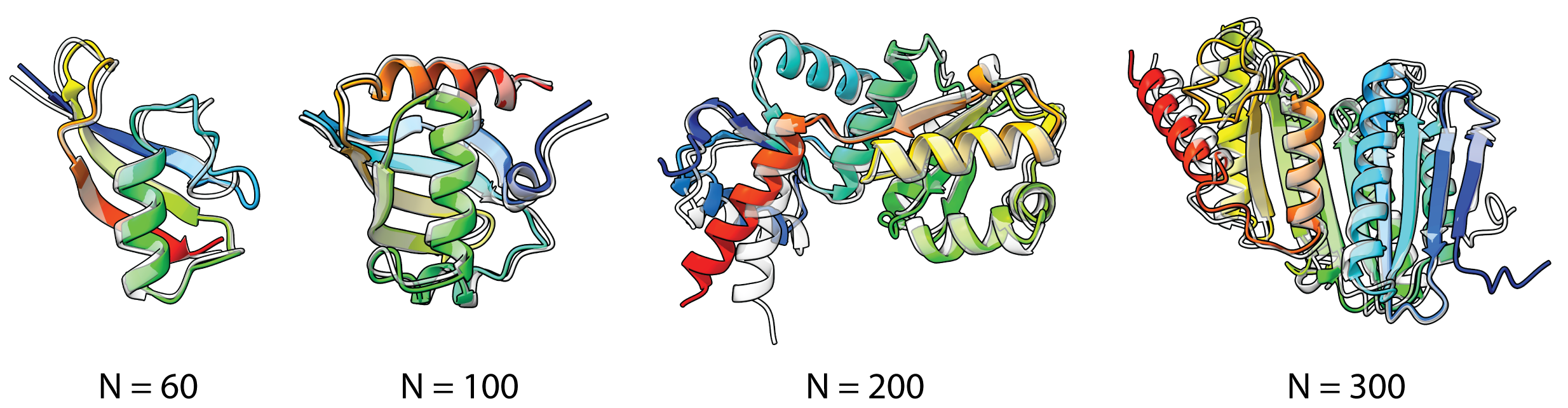}
    \caption{Representative examples of designable protein backbones generated with GAFL (white) and the output of the refolding pipeline (colored), comprising ProteinMPNN and ESMFold.\vspace{-0.15cm}
    }
    \label{fig:designs_GAFL}
\end{figure}

We train GAFL\footnote{Source code and trained model weights are available at \url{https://github.com/hits-mli/gafl}} on a subset of the Protein Data Bank (PDB) dataset~\citep{berman2000protein} comprised of monomeric protein structures with up to 512 residues and perform extensive ablations on the smaller, curated SCOPe dataset~\cite{fox2014scope, chandonia2022scope} filtered by proteins with length of up to 128 residues (SCOPe-128) as in~\citep{yim2023frameflow, lin2023genie}.
A representative selection of designable protein backbones generated by GAFL trained on the PDB dataset is illustrated in Figure~\ref{fig:designs_GAFL}.

\subsection{Checkpoint Selection} \label{sec:training_details}

With the aim of finding a good balance between designability and secondary structure content in mind, we introduce a \emph{checkpointing criterion} that takes secondary structure into account.
We first train the model for $N_\text{train}$ epochs.
Then, from epoch $N_\text{train}$ to $N_\text{train}+N_\text{select}$, we calculate the relative occurrence of $\alpha$-helices and $\beta$-strands $r_\alpha$ and $r_\beta$ of 100 generated proteins after each epoch.
We keep the top $k$ checkpoints in terms of secondary structure content deviation, which we define as $d_c \equiv |r_\alpha-r_\alpha^{(\text{ref})}| + |r_\beta-r_\beta^{(\text{ref})}|\,$, with the training set as reference.
Among those $k$ checkpoints, we choose the checkpoint with the highest designability and filter by a threshold $d_c \leq d_\text{max}$.
The hyperparameters $N_\text{train}$, $N_\text{select}$, $d_\text{max}$ and $k$ for the different training runs are listed in Appendix~\ref{sec:ckpt_hyperparams}.

\subsection{Metrics} \label{sec:metrics}
To assess the performance of a given model, we follow a well-established pipeline of self-consistency evaluation~\citep{trippe2022diffusion, watson2023novo}.
For each generated backbone, we design 8 candidate sequences with ProteinMPNN~\cite{dauparas2022protmpnn}, which are subsequently refolded with ESMfold~\citep{lin2023evolutionary}, and define $\text{scRMSD}$ as the smallest RMSD between our generated backbone and the 8 refolded, aligned candidates.
As in~\cite{watson2023novo, yim2023framediff}, we define \emph{designability} as the fraction of generated samples with $\text{scRMSD}$ < 2.0 \AA \,.
We also report \emph{diversity} and \emph{novelty} of the designable backbones as average TM-similarity~\citep{zhang2004scoring} scores within the set of generated backbones and with respect to the PDB correspondingly (see Appendix~\ref{sec:novely_definition} and~\citep{yim2023frameflow}).
To evaluate how well the secondary structure distribution of the training set is captured, we calculate the \emph{average helix and strand content} of all designable backbones using the DSSP algorithm~\citep{kabsch1983dictionary}.

\subsection{Baselines} \label{sec:baselines_training}

At the task of generating backbones of up to 300 residues, we compare GAFL trained on the PDB to the diffusion models RFdiffusion~\citep{watson2023novo} and FrameDiff~\citep{yim2023framediff} and to the flow matching models FoldFlow~\citep{bose2023se} and FrameFlow~\citep{yim2023frameflow, yim2024improved}.
At the time of submitting the paper, FrameFlow had not been trained on the PDB yet, however, we include it in our results as contemporary work.
Further, we perform an ablation study for generating smaller backbones, where we compare GAFL models to the originally published FrameFlow~\citep{yim2023frameflow} model, which was trained on the SCOPe dataset with backbones of up to 128 residues.
While all of the baselines above incorporate the original IPA~\cite{jumper2021highly} architecture, we also compare GAFL with VFN~\cite{mao2023modeling}, in which an alternative modification of IPA is proposed.
For all models considered, we generate backbones using published model weights and the respective default inference settings (Appendix \ref{sec:baselines}).

\subsection{Results}
\label{sec:results}

We train GAFL for 15 days on two NVIDIA A100-80GB GPUs on the dataset used in FrameDiff, which comprises monomeric structures from the Protein Data Bank (PDB), filtered by a maximum length of 512 and a maximum coil content of 50\%, resulting in a total of around 25,000 backbones \cite{yim2023framediff}.
As in FoldFlow, we evaluate GAFL and the respective baselines on the task of generating backbones of lengths $\{100,150,200,250,300\}$.
For GAFL, we use 200 inference timesteps (Figure~\ref{fig:timesteps_GAFL}).
In Table~\ref{tab:pdb-performance}, we report the metrics described in Section~\ref{sec:metrics} for each model together with the time needed to generate a single backbone of length 100 on an NVIDIA A100 GPU without batching.

\begin{figure}[t]
    \centering
    \includegraphics[width=1.0\linewidth]{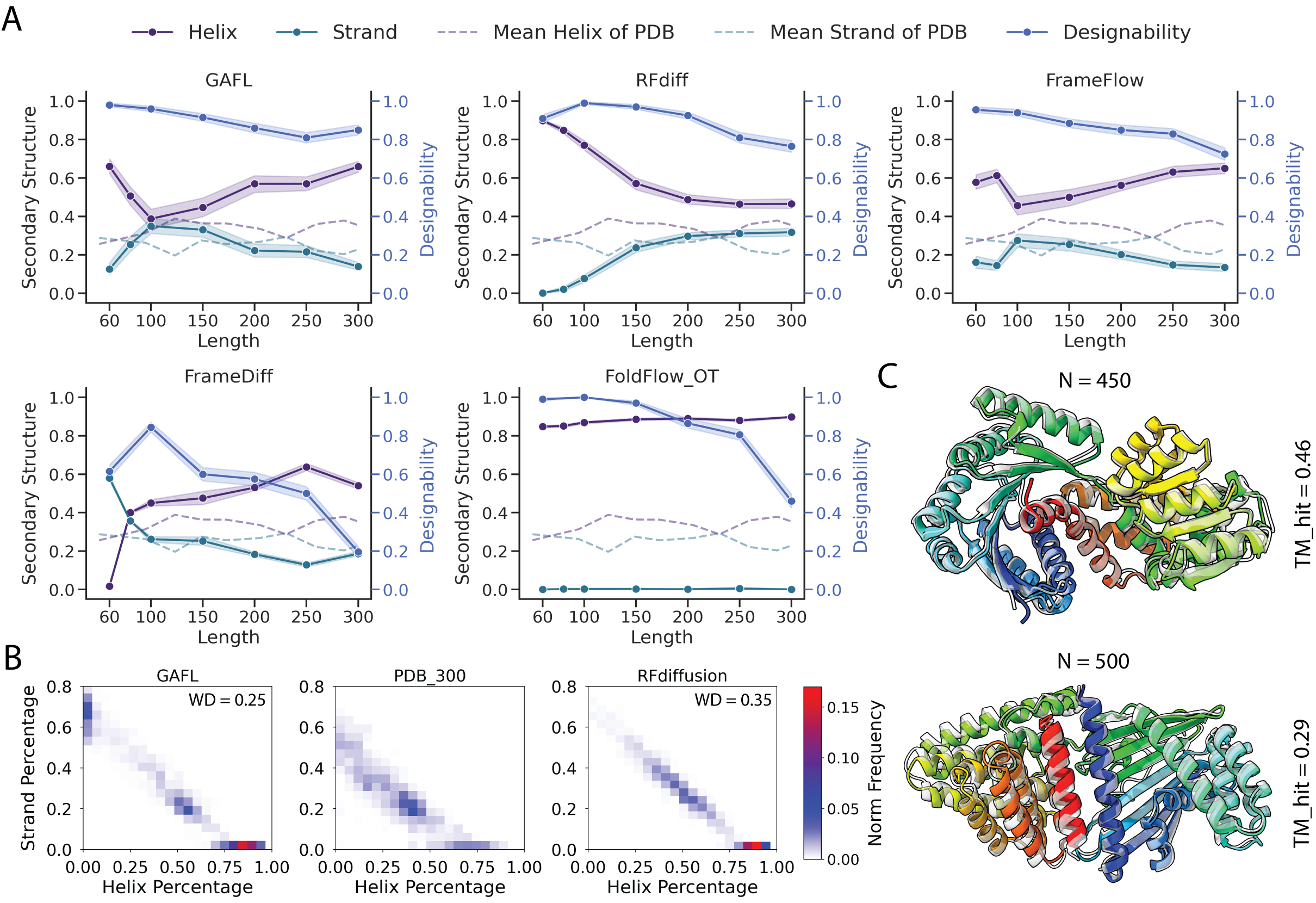}
    \caption{(\textbf{A}) Performance of evaluated models in terms of designability and secondary structure content as a function of backbone length. 200 backbones were generated for each model at each length $\in\{60,80,100,150,200,250,300\}$. (\textbf{B}) Comparison of the secondary structure distributions of backbones generated by GAFL and RFdiffusion from (A) to the PDB dataset filtered by the respective protein lengths along with the Wasserstein distance (WD) between the distributions. (\textbf{C}) Examples of designable backbones generated by GAFL for lengths 450 and 500.
    We also report TM scores of the backbones to the closest hit in the PDB database computed with FoldSeek.\vspace{-0.15cm}}
    \label{fig:des_by_len_combined}
\end{figure}

\begin{table}[]
\small
    \centering
    \caption{
    Performance of GAFL and baseline models for the generation of 200 protein backbones for each length in $\{100, 150, 200, 250, 300\}$.
    We report the metrics from Section~\ref{sec:metrics}, including standard errors obtained by bootstrapping, and the time needed to generate a backbone of length 100.
    The best values and values within the respective margin of error are bold.
    }
    \resizebox{\textwidth}{!}{%
    \begin{tabular}{@{}lcccccc@{}}
    \toprule
    Method           & Designability ($\uparrow$)  & Diversity ($\downarrow$) & Novelty ($\downarrow$) & Helix Content & Strand Content & Time [s]\\
    \midrule
    PDB Dataset (300)      &  - &   -   & -   & 0.39 (0.00)      & 0.23 (0.00)     &      \\
    \midrule
    FrameDiff               & 0.54 (0.02)          & 0.45 (0.00) & 0.71 (0.00)     & \textbf{0.53} (0.01)     & 0.20 (0.00)      &   24.3  \\
    FoldFlow-SFM       & 0.69 (0.01)        & 0.44 (0.00)     & 0.77 (0.00)    & 0.91 (0.00)      & 0.01 (0.00)      &    24.3   \\
    FoldFlow-OT             & 0.82 (0.01) & 0.44 (0.00)          & 0.79 (0.00)      & 0.88 (0.00)      & 0.00 (0.00)        &   24.3   \\
    FrameFlow                 & 0.85 (0.01)  & \textbf{0.35} (0.00)              & \textbf{0.70} (0.00)     & 0.56 (0.01)       & 0.20 (0.00)   &  6.6 \\
    RFdiffusion${^*}$  & \textbf{0.89} (0.01)         & 0.37 (0.00)            & 0.74 (0.00)     & 0.58 (0.02)     & \textbf{0.24} (0.02)    &   21.0   \\
    \midrule
    GAFL (ours)                   & \textbf{0.88} (0.01)  & 0.36 (0.00)              & 0.71 (0.00)      & \textbf{0.53} (0.01)       & \textbf{0.25} (0.01)   &   8.8 \\
    \bottomrule
    \multicolumn{6}{l}{\footnotesize{*Pretrained weights from folding model trained on dataset larger than PDB.}}
    \end{tabular}%
    }
    \vspace{-0.23cm}
    \label{tab:pdb-performance}
\end{table}

\subsubsection*{GAFL has state-of-the-art performance}
We find that GAFL can reliably generate designable, diverse and novel backbones while capturing the statistical distribution of secondary structure elements of natural proteins.
In all metrics considered, GAFL outperforms both variants of FoldFlow and is better or as good as FrameDiff.
GAFL also outperforms FrameFlow, which was trained on the same dataset, in terms of designability and, at the same time, achieves better secondary structure content.
GAFL's designability is only matched by RFdiffusion, which is not directly comparable since it relies on pre-trained model weights from the folding model RoseTTAFold~\citep{baek2021accurate} and has around three times more parameters.
For diversity, novelty and helix content, GAFL performs better than RFdiffusion.
We also observe that GAFL and FrameFlow can generate backbones around three times faster than the other evaluated models.

\subsubsection*{GAFL generates proteins with diverse secondary structures at various lengths}
Although protein design campaigns span a wide range of protein sizes~\citep{torresnovo, cao2020novo, olshefsky2022engineering, bennett2023improving, watson2023novo}, the primary goal remains to encode maximum functionality into the smallest protein possible, driven by the growing costs of synthesis as protein size increases.
Thus, we further assess designability and secondary structure content of generated backbones as a function of their length (Figure~\ref{fig:des_by_len_combined}A).
For all lengths considered, backbones generated by GAFL, FrameFlow and RFdiffusion are highly designable while FrameDiff and FoldFlow struggle with the generation of long proteins.
The length dependence of designability and secondary structure content for GAFL and FrameFlow is qualitatively similar; however, GAFL achieves overall better results~(Table~\ref{tab:pdb-performance}, Figure~\ref{fig:gafl_ff_des_by_len}).
Crucially, we find that for generating proteins with less than 150 residues, GAFL is well suited as it is capable of generating highly designable backbones with a similar amount of $\beta$-strands as naturally occurring proteins, while RFdiffusion over-represents $\alpha$-helices.
This is also reflected by the Wasserstein distances between the secondary structure distribution of naturally occurring backbones and those generated by GAFL and RFdiffusion, respectively:
Purely $\alpha$-helical proteins are over-represented by RFdiffusion, leading to a higher Wasserstein distance of 0.35 compared to 0.25 for GAFL (Figure \ref{fig:des_by_len_combined}B).

\subsubsection*{GAFL can generate large proteins}
To compare GAFL with VFN~\cite{mao2023modeling}, we evaluate it at generating five backbones for each length in $\{100,105,\ldots,500\}$, some of which are portrayed in Figure \ref{fig:des_by_len_combined}C.
We find that, with a value of 0.74, GAFL outperforms not only VFN (0.44) but also FrameFlow (0.64) and RFdiffusion (0.71) in designability.
However, GAFL and FrameFlow over-represent helices for large proteins (Table \ref{tab:GAFL_VFN_long}).

\subsection{Ablation of CFA} \label{sec:ablations}

In order to investigate the effect of the proposed architectural changes, we conduct an ablation study on the semi-manually curated SCOPe dataset~\cite{fox2014scope, chandonia2022scope},
which clusters proteins by their sequence and structural similarities ensuring that its entries are evolutionary and structurally non-redundant.
We further filter SCOPe by the length of up to 128 residues, which results in 3938 proteins.

\begin{wrapfigure}{r}{0.5\textwidth}
  \centering
  \includegraphics[width=0.45\textwidth]{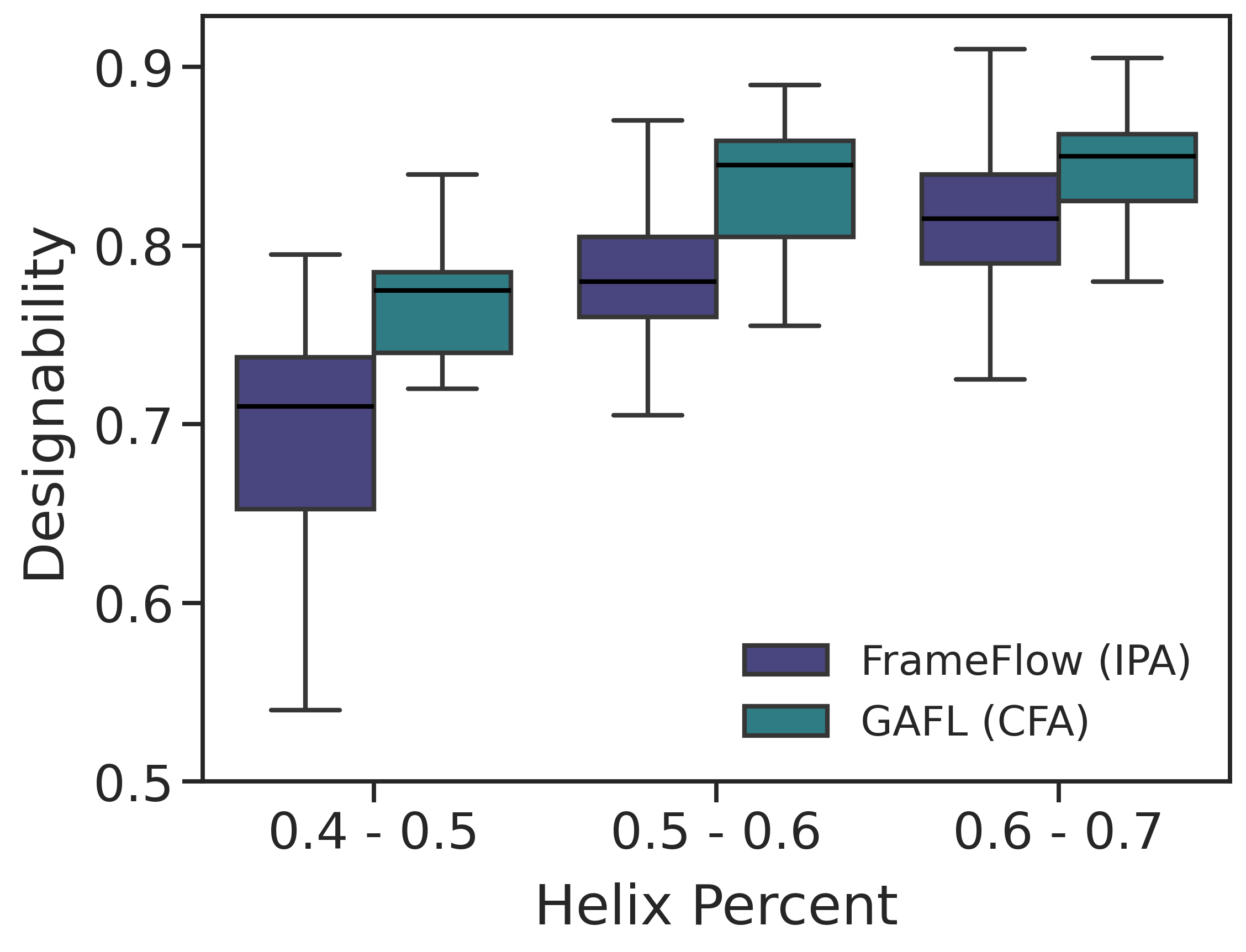}
  \vspace{-0.2cm}
  \caption{Helix content and designabilities of 90 model checkpoints sampled during three training runs on the PDB dataset for GAFL and retrained FrameFlow, respectively. For each checkpoint, we sample 40 backbones per length in $\{100,150,\ldots,300\}$.}
  \label{fig:GAFL_FF_boxplot_helices}
\end{wrapfigure}

We compare GAFL with FrameFlow and models for which we leave out all proposed architectural changes or only higher order message passing, respectively, while scaling the width of the layers such that the number of parameters remains the same.
All models are trained with three different random seeds for 6500 epochs on one NVIDIA A100-80GB GPU, which takes around 6 days, and evaluated by generating backbones for lengths between 60 and 128 as in~\cite{yim2023frameflow}.

\begin{wraptable}{r}{0.5\textwidth}
\small
    \centering
    \caption{Ablation of GAFL and FrameFlow on the PDB dataset. We report the results of three training runs and the published FrameFlow model.
}
\begin{tabular}{lcc}
\toprule
Method           & Designability ($\uparrow$)  & Diversity ($\downarrow$)\\
\midrule
GAFL                   & $\error{87.8}{87.9}{84.3}$  & $\error{0.35}{0.36}{0.34}$\\
FrameFlow         & $\error{84.4}{84.7}{78.1}$  & $\error{0.35}{0.35}{0.34}$   \\ 
FrameFlow$^*$ & 84.6 & 0.35\\
\bottomrule
\multicolumn{2}{l}{\footnotesize{*Published model weights}}
\end{tabular}%
\label{tab:gafl_ff_PDB}
\end{wraptable}

We find that GAFL's designability of 90.5\% is 8 percentage points higher than that of the published FrameFlow model~\cite{yim2023frameflow} trained on SCOPe (81.2\%) as depicted in Table~\ref{tab:ablation}.
If retrained with GAFL's training procedure, FrameFlow's architecture with original IPA achieves a designability of 88.2\%.
Using PGA-valued features and higher order message passing as proposed in CFA increases the designability by 1.4 and 2.3 percentage points, respectively.
While the training procedure has a larger effect on the designability, the improvement due to CFA can be considered to be significant since it persists across different random seeds.
This can also be observed in the distribution of designabilities of 30 checkpoints sampled during the training procedure (Figure \ref{fig:scope_training}) described in Section~\ref{sec:training_details}.

For the much larger PDB dataset, we conduct a small ablation study, in which we compare the performance of GAFL with CFA and retrained FrameFlow with original IPA across three different training runs, respectively (Table~\ref{tab:gafl_ff_PDB}).
Since GAFL consistently achieves higher designabilities, we can validate that the trend observed on SCOPe also holds true for training on the PDB.

Furthermore, we observe a correlation between over-representing $\alpha$-helices and achieving high designability (Figure~\ref{fig:GAFL_FF_boxplot_helices}) for 90 checkpoints sampled for each model during checkpoint selection as described in Section~\ref{sec:training_details}.
Crucially, GAFL checkpoints consistently show better designability in the low helix-content regime, which we can attribute to using CFA instead of IPA.

\begin{table}
\centering
\caption{Ablation of GAFL models with different elements of the proposed CFA architecture, trained on the SCOPe-128 dataset.
For each model, we report the performance of three training runs with different random seeds, evaluated by generating 10 backbones for each length in $\{60,61,\ldots,128\}$.
}

\renewcommand{\arraystretch}{1.3}  
\begin{tabular}{@{}ccc|ccc@{}}
\toprule
PGA & Higher Ord. Msg. & Checkp. Selection & Designability [\%] ($\uparrow$) & Diversity ($\downarrow$) \\
\midrule
\checkmark                 &    \checkmark     & \checkmark    &       $\error{90.5}{90.6}{89.6}$      &     $\error{0.38}{0.39}{0.36}$               \\
\checkmark                 &         & \checkmark      &            $\error{89.6}{90.3}{89.0}$  &  $\error{0.37}{0.40}{0.37}$             \\
                 &          & \checkmark    &      $\error{88.2}{88.6}{86.7}$        &      $\error{0.38}{0.39}{0.38}$              \\
\hline
\multicolumn{2}{l}{FrameFlow-SCOPe}          &     &    81.2        &      0.37                  \\
\bottomrule
\end{tabular}%
\label{tab:ablation}
\vspace{-0.05cm}
\end{table}

\subsection{Discussion}
Our results suggest that GAFL is one of the current state-of-the-art models for unconditional protein structure generation.
GAFL outperforms the widely used, pre-trained model RFdiffusion in terms of diversity, novelty and inference time and is on par at designability.
Remarkably, GAFL can achieve this performance without requiring pre-trained model weights while other non-pre-trained models often lack designability or show a mode-collapse towards generating helical structures.

Especially for generating small, highly designable backbones with distinct secondary structures, GAFL performs well, in particular better than RFdiffusion.
Since in most protein design campaigns the goal is to incorporate the desired functionality into the smallest protein possible while exploring a large structural space, we consider this advantage of GAFL to be highly relevant for future developments and real-world applications of generative models for proteins.

Our ablation studies provide evidence that achieving high designability without over-representing $\alpha$-helices can be attributed to the replacement of IPA by the proposed CFA architecture.
Since IPA is used in many current state-of-the-art architectures for backbone structure, CFA has the potential to enable improvements in many protein-related tasks.

\paragraph{Limitations}
While the proposed method GAFL achieves state-of-the-art performance in protein backbone generation, there is still room for improvement.
We note that achieving high designability without compromising the diversity of secondary structures remains a grand challenge.
GAFL does not perfectly capture the secondary structure distribution of natural proteins, especially for large proteins.
Further, unconditional protein generation can be regarded as suitable benchmarking task for protein design models, however, most applications require conditional sampling.
GAFL can be readily incorporated into existing frameworks for conditioning, e.g. on motifs~\cite{yim2024improved} or symmetry~\cite{watson2023novo}.

%% file: 5_conclusion.tex
\section{Conclusion}
We introduced Geometric Algebra Flow Matching (GAFL), a flow matching model for protein design based on FrameFlow~\citep{yim2023frameflow}.
GAFL relies on the proposed Clifford Frame Attention (CFA), an extension of the invariant point attention block from AlphaFold2~\cite{jumper2021highly}, by representing residue frames and geometric features of a protein in the projective geometric algebra.
This enables the usage of the bilinear operations in the algebra to construct geometrically expressive messages between residues. 
The experiments demonstrate that the resulting model is state-of-the-art in the combination of the established metrics designability, diversity and novelty and performs notably well at generating designable small backbones with distinct secondary structures, containing $\beta$-strands in particular.
Given the promising results of the extension of invariant point attention with geometric algebra presented in this work, we look forward to exploring its benefits for other protein-related tasks.

\paragraph{Acknowledgments} This study received funding from the Klaus Tschira Stiftung gGmbH (HITS Lab). We acknowledge the National Academic Infrastructure for Supercomputing in Sweden (NAISS), partially funded by the Swedish Research Council through grant agreement no. 2021-29 for awarding this project access to the Berzelius resource provided by the Knut and Alice Wallenberg Foundation at the National Supercomputer Centre. The authors acknowledge support by the state of Baden-Württemberg through bwHPC and the German Research Foundation (DFG) through grant INST 35/1597-1 FUGG.

%% file: 6_appendices.tex
\section{Appendix}

\setcounter{table}{0}
\renewcommand{\thetable}{\thesection.\arabic{table}}
\renewcommand*{\theHtable}{\thetable}

\subsection{Background}

\subsubsection{Background on Geometric Algebra}
\label{sec:ga_background}

In this section we give a short introduction to \emph{Clifford algebra}, loosely following the presentation of \citet{dorst2007geometric}, \citet{dorst2020guided} and \citet{doran2003geometric}.

\paragraph{General construction of Clifford algebras}\mbox{}\\

A Clifford algebra can be constructed from a vector space $V$ by extending it with an additional bilinear operation called the \emph{geometric product}. We write this algebra as $\cliff(V)$. Elements of the algebra  $\mvec{A} \in \cliff(V)$ are called \emph{multivectors} and are written in bold. In cases where we want to highlight that an algebra element coincides with a vector in $V$, we write it with a (bold) lower case letter. Other algebra elements are generally denoted by upper case letters. The geometric product has to fulfill the following properties \cite{doran2003geometric}:

\begin{mathdef}{Properties of the geometric product}
    \deflabel{gp_properties}

\begin{enumerate}
    \item Associativity: $(\mvec{A} \mvec{B}) \mvec{C} = \mvec{A} (\mvec{B} \mvec{C}) \hfill \mvec{A}, \mvec{B}, \mvec{C} \in \cliff(V)$
    \item Distributivity: $\mvec{A}(\mvec{B} + \mvec{C}) = \mvec{A} \mvec{B} + \mvec{A} \mvec{C}$
    \item Vectors square to scalars: $\mvec{a} \mvec{a} = \mvec{a}^2 \in \mathbb{R}$
\end{enumerate}

\end{mathdef}

We denote the geometric product as juxtaposition of elements in order to distinguish it from other products liker inner and outer product. 

\subparagraph*{Algebra basis}

Given a basis of the underlying vector space $V$, $\{\mvec{e}_i\}_{i=0}^n$, where $n$ is the dimension of the vector space, we can use the geometric product to construct a basis for the algebra.

\begin{equation*}
    \{1, \mvec{e}_1, \dots, \mvec{e}_n, \mvec{e}_1\mvec{e}_2, \mvec{e}_1\mvec{e}_3, \dots, \mvec{e}_1 \dots \mvec{e}_n\}
\end{equation*}

In general, a geometric algebra over an $n$ dimensional vector space will have $2^n$ basis vectors. A general multivector can be written as a linear combination of these basis vectors.
The basis elements can be further categorized into so called \emph{grades} according to the number of basis vectors they contain, i.e. $1$ will be of grade 0, $\mvec{e}_i$ of grade 1, $\mvec{e}_i \mvec{e}_j$ of grade 2 and so on.
With this definition it is possible to define the projection of an arbitrary multivector onto a specific grade \cite{dorst2007geometric}.

\begin{mathdef}{Grade projection}
\deflabel{grade_projection}

    Let $\cliff^{[k]}(V)$ be the subspace of $\cliff(V)$ spanned by all multivectors of grade $k$. We then define the grade projection operator
    \begin{equation}
    \eqlabel{grade_projection}
        \grade{\cdot}{k}: \cliff(V) \xrightarrow{} \cliff^{[k]}(V)\,,
    \end{equation}
    which selects the $k$-th grade of a given multivector.
\end{mathdef}

Multivectors that contain only elements of one grade receive specific names according to their grade. Elements of grade 1 are \emph{vectors}, elements of grade 2 are \emph{bivectors}, elements of grade 3 are \emph{trivectors} and so on.

\subparagraph*{The metric}

So far we have only specified that the square of a vector should yield a scalar. In order to uniquely define a geometric algebra it is important to specify the exact values of these scalars, which will define a metric on the algebra. The choice of metric is crucial for the properties of the algebra and the geometric interpretation of its elements. For an algebra over an $n$-dimensional vector space we may choose $n$ scalars, one for each basis vector. It is common to work with a metric that assigns 1 or -1 to all non-zero values. This convention leaves $\{-1,0,1\}$ as possible choices for the scalars. One defines $\cliff{n,m,l}$ as the geometric algebra with $n$ basis vectors squaring to $1$, $m$ basis vectors squaring to $-1$ and $l$ basis vectors squaring to $0$.
Having fixed a metric for the vectors of the algebra we can go on to discuss the construction of a norm for general multivectors. The definition of the norm in geometric algebra uses the concept of \emph{reversion} which is defined as follows:

\begin{mathdef}{Reversion}
    \deflabel{reverse}

    Let $\mvec{a}_1, \mvec{a}_2, \dots \mvec{a}_n \in \cliff^{[1]}(V), \; n \in \mathbb{N}$ be vectors and $\mvec{A} = \prod\limits_{i=1}^n\mvec{a}_i$ the geometric product of these vectors, then the reverse of $\mvec{A}$ is defined as
    \begin{equation}
        \tilde{\mvec{A}} = \prod\limits_{i=1}^n\mvec{a}_{n-i+1},
        \label{eq:def_reverse}
    \end{equation}
    i.e. we reverse the order of the vector elements. For a general multivector, which may consist of a sum of elements of the form of \ref{eq:def_reverse}, we apply the reverse operation to each summand individually.

\end{mathdef}

A norm on general mutltivectors can then be defined as:

\begin{mathdef}{Norm}

    \deflabel{ga_norm}

    Let $\mvec{A} \in \cliff(V)$ be a multivector, then the norm of $\mvec{A}$ is defined as

    \begin{equation}
        \eqlabel{ga_norm}
        \| \mvec{A} \| = \sqrt{\langle \tilde{\mvec{A}} \mvec{A} \rangle_0}.
    \end{equation}
    
\end{mathdef}

\paragraph{The Euclidean geometric algebra \texorpdfstring{$\cliff{3}$}{Cl(3)}}\mbox{}\\
\seclabel{ega}

One of the most prominent examples for a geometric algebra is the Euclidean geometric algebra $\cliff{3}$, which is the geometric algebra over $\mathbb{R}^3$ with the standard Euclidean metric, i.e. a vector basis $\{\mvec{e}_1, \mvec{e}_2, \mvec{e}_3\}$ fulfilling $\mvec{e}_1^2 = \mvec{e}_2^2 = \mvec{e}_3^2 = 1$. Apart from the geometric product we can define two additional operations that will be useful for further analysis.

\begin{mathdef}{Inner and outer product}
    \deflabel{inner_outer_product}

    Given two arbitrary vectors $\mvec{a}, \mvec{b} \in \cliff{3}^{[1]}$ we can separate their geometric product into a symmetric and antisymmetric part
    \begin{equation}
        \mvec{a} \mvec{b} = \frac{1}{2}\left(\mvec{a} \mvec{b} + \mvec{b} \mvec{a}\right) + \frac{1}{2}\left(\mvec{a} \mvec{b} - \mvec{b} \mvec{a}\right).
    \end{equation}
    One typically refers to $\frac{1}{2}\left(\mvec{a} \mvec{b} + \mvec{b} \mvec{a}\right) =: \mvec{a} \cdot \mvec{b}$ as the \defemph{inner product} and $\frac{1}{2}\left(\mvec{a} \mvec{b} - \mvec{b} \mvec{a}\right) =: \mvec{a} \wedge \mvec{b}$ as the \defemph{outer product} of the two vectors. The geometric product of $\mvec{a}$ and $\mvec{b}$ can then be written as
    \begin{equation}
        \eqlabel{gp_vector_decomposition}
        \mvec{a} \mvec{b} = \mvec{a} \cdot \mvec{b} + \mvec{a} \wedge \mvec{b}.
    \end{equation}
\end{mathdef}

This definition only holds for vectors. For an extension to arbitrary multivectors see \cite{doran2003geometric}. To further investigate the properties of these products we can look at the square of the sum of two vectors $\left(\mvec{a} + \mvec{b}\right)^2$. Expanding the product and rearranging the terms yields

\begin{equation}
    \eqlabel{polarization}
    \frac{1}{2} \left(\mvec{a} \mvec{b} + \mvec{b} \mvec{a}\right) = \frac{1}{2}\left(\mvec{a} + \mvec{b}\right)^2 - \mvec{a}^2  - \mvec{b}^2 = \mvec{a} \cdot \mvec{b}.
\end{equation}

Since all terms on the right side are scalars due to the third property in \defref{gp_properties}, we can conclude that the inner product of two vectors is a scalar itself. In fact, \eqref{polarization} corresponds to the well known polarization identity from linear algebra, and since we defined the norm of vectors to be the standard Euclidean norm, the inner product from \defref{inner_outer_product} is the standard Euclidean inner product.

Using \eqref{gp_vector_decomposition} we can also introduce the notion of an orthogonal basis. Just as in linear algebra a vector basis $\{\mvec{e}_i\}_i$ is called \emph{orthogonal} if $\mvec{e}_i \cdot \mvec{e}_j = 0$ for $i \neq j$. In the following discussion we will always assume an orthogonal basis. As a consequence of \eqref{gp_vector_decomposition} for $i \neq j$ basis elements anticommute $\mvec{e}_i \mvec{e}_j = - \mvec{e}_j \mvec{e}_i$ and the geometric product is equal to the outer product $\mvec{e}_i \mvec{e}_j = \mvec{e}_i \wedge \mvec{e}_j$.

\subparagraph*{Geometric interpretation of the algebra elements}

\begin{figure}[t]
    \centering
    \includegraphics[width=0.9\textwidth]{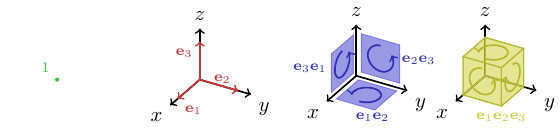}
    \caption{Visualization of the different geometric primitives in $\cliff{3}$. Vectors are directed line segments, bivectors are oriented areas and trivectors are oriented volumes. Orientations are indicated by arrows that have the same sense of rotation as the corresponding basis vectors when linked together end to tip.}
    \label{fig:ega_primitives}
\end{figure}

As the name \emph{geometric} algebra suggests, it is possible to assign geometric meaning to the algebra elements with grade greater than 0. Grade one elements or \emph{vectors} keep their usual geometric interpretation as directed line segments. To find an interpretation for higher grade elements, one can observe that the outer product of two vectors $\mvec{A} = \mvec{a} \wedge \mvec{b}$ defines a homogeneous subspace in the sense that for every vector in the span of $\mvec{a}$ and $\mvec{b}$ the following equation holds

\begin{equation}
    \eqlabel{subspace_rep}
    \mvec{x} \in \mathcal{A} = \spn{\{\mvec{a}, \mvec{b}\}} \iff \mvec{x} \wedge \mvec{A} = 0,
\end{equation}
as shown in \cite{dorst2007geometric}. Additionally $\mvec{A}$ has a magnitude according to \defref{ga_norm} and also an orientation due to the antisymmetry of the outer product. For example $\mvec{e}_1 \wedge \mvec{e}_2$ has opposite orientation compared to $\mvec{e}_2 \wedge \mvec{e}_1 = - \mvec{e}_1 \wedge \mvec{e}_2$ as indicated by the relative minus sign. 2-\emph{blades}, that is multivectors which can be written purely as the outer product of two vectors, can thus be interpreted as oriented areas, which lie within the subspace spanned by their generating vectors and area equal to the norm of the blade. Analogously 3-blades, multivectors equal to the outer product of three vectors, correspond to oriented volumes. Notably like vectors, 2-blades and 3-blades neither have a position in space nor do they have a specified shape. The only fixed properties are the magnitude of their area/volume as well as their orientation. The geometric interpretation of algebra elements is visualized in \figref{ega_primitives}.

Grades of the same dimensionality are closely related by a relation called \emph{duality}. Graphically speaking, a plane in 3D space for example can be represented both as the plane itself or by the normal vector that is orthogonal to it. In algebraic terms this translates to e.g. $\mvec{e}_3$ representing the normal vector of the plane described by the bivector $\mvec{e}_1\mvec{e}_2$. One says that these two elements are \emph{dual} to each other. In the same way, scalars are dual to trivectors since trivectors, like scalars, have only one degree of freedom in three dimensions and are thus also sometimes called pseudoscalars. Formally duality in $\cliff{3}$ can be defined as follows:

\begin{mathdef}{Duality}
    \deflabel{ega_duality}

    Let $\mvec{A} \in \cliff{3}$ and $\egaI = \mvec{e}_1\mvec{e}_2\mvec{e}_3$. Then the dual of $\mvec{A}$ is given by

    \begin{equation}
        \eqlabel{ega_duality}
        \egadual{\mvec{A}} = \egaI \mvec{A}
    \end{equation}

\end{mathdef}

Finally, it is also insightful to calculate the magnitude of the area represented by a 2-blade. To this end we notice that, using the Einstein sum convention, the outer product of two vectors $\mvec{a} = a_i \mvec{e}_i, \, \mvec{b} = a_j \mvec{e}_j$ can be written as

\begin{align}
    \mvec{a} \wedge \mvec{b} =  a_i a_j \mvec{e}_i \wedge \mvec{e}_j = a_i a_j \mvec{e}_i \mvec{e}_j = \epsilon_{ijk} a_i b_j \egaI \mvec{e}_k = \egaI \left( \mvec{a} \times \mvec{b} \right),
\end{align}

where $\epsilon_{ijk}$ is the Levi-Civita symbol and $\times$ denotes the usual vector cross product. This shows that in $\cliff{3}$ the outer product yields the 2-blade which is dual to the resulting vector of the cross product with the same magnitude. In fact, the outer product can be seen as generalization of the cross product to arbitrary dimensions. The magnitude can be calculated from \eqref{ga_norm}, yielding

\begin{equation}
    \|\mvec{a} \wedge \mvec{b}\| = \sqrt{\langle \| \mvec{a} \times \mvec{b} \|^2 \tilde{\egaI} \egaI \rangle_0} = \sqrt{ \| \mvec{a} \times \mvec{b} \|^2} = \|\mvec{a}\| \|\mvec{b}\| |\sin(\alpha)|,
\end{equation}

where $\alpha$ is the angle enclosed by $\mvec{a}$ and $\mvec{b}$. This is exactly the area of the parallelogram spanned by the two vectors. Similar calculations can be performed for the case of 3-blades to show that their magnitude is equal to the volume of the parallelepiped spanned by their three generating vectors. These findings further support the idea of interpreting bivectors and trivectors as area and volume elements respectively.

\subparagraph*{Orthogonal transformations}

Another important class of operations which geometric algebra can describe in an efficient way are orthogonal transformations, that is transformations $T: \cliff{3} \rightarrow \cliff{3}$ which preserve the inner product between vectors. In group theoretic terms, these transformations form the group $\mathrm{O}(3)$. In order to construct such transformations we first look at how reflections are handled in $\cliff{3}$.

\begin{mathdef}{Reflection}

    Let $\mvec{a} \in \cliff{3}^{[1]}$ be a vector and $\mvec{n} \in \cliff{3}^{[1]}$ be the unit normal vector of a plane. The reflection of $\mvec{a}$ at that plane is then given by

    \begin{equation}
        \mvec{a}' = -\mvec{n}\mvec{a}\mvec{n}.
    \end{equation}

\end{mathdef}

To see why this indeed corresponds to a reflection, we decompose $\mvec{a}$ into a parallel and orthogonal part with respect to the normal vector $\mvec{n}$ as shown in \figref{reflection}. Algebraically this can be written as

\begin{figure}[t]
    \centering
    \includegraphics[width=0.3\textwidth]{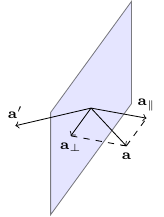}
    \caption{Visualization of the reflection of a vector $\mvec{a}$ at a plane.}
    \figlabel{reflection}
\end{figure}

\begin{align}
\mvec{a} &= \mvec{n}^2 \mvec{a} = \mvec{n}(\mvec{n}\mvec{a}) = \mvec{n}(\mvec{n}\cdot \mvec{a} + \mvec{n} \wedge \mvec{a}) \\
&= \underbrace{\mvec{n} (\mvec{n} \cdot \mvec{a})}_{=\mvec{a}_{\parallel}} + \underbrace{\mvec{n} (\mvec{n} \wedge \mvec{a})}_{=\mvec{a}_{\perp}}.
\end{align}

The first summand corresponds to the usual projection formula from linear algebra and can thus be identified as the parallel part. Since the whole expression equals $\mvec{a}$, the second summand must be the orthogonal part. The reflected vector can now be obtained by reversing the sign of $\mvec{a}_{\parallel}$.

\begin{align}
    \eqlabel{reflection_deriv1}
    \mvec{a}' &= -\mvec{a}_{\parallel} + \mvec{a}_{\perp} = -\mvec{n}(\mvec{a}\cdot \mvec{n}) + \mvec{n}(\mvec{n}\wedge \mvec{a})\\ 
    \eqlabel{reflection_deriv2}
    &= -\mvec{n}(\mvec{a}\cdot \mvec{n}) - \mvec{n}(\mvec{a}\wedge \mvec{n}) = - \mvec{n}(\mvec{a}\cdot \mvec{n} + \mvec{a}\wedge \mvec{n})\\
    &= -\mvec{n}\mvec{a}\mvec{n}.
\end{align}

From \eqref{reflection_deriv1} to \eqref{reflection_deriv2}, we used the antisymmetry of the outer product. This construction of reflections shows that elements of Euclidean geometric algebra can simultaneously be interpreted as geometric objects as well as transformation operators, a concept which also translates to other geometric algebras.

To construct general orthogonal transformations we can make use of the following theorem. 

\begin{theorem}{Cartan-Dieudonné theorem}
    \thmlabel{cartan_dieudonne}

    Let $(V, q)$ be a nondegenerate space of dimension $n$, then any orthogonal transformation can be written as a composition of at most $n$ reflections.

\end{theorem}

For a proof see \cite{lundholm2009clifford}. In the context of $\cliff{3}$, this implies that any orthogonal transformation of a vector, which in addition to reflections comprises rotations around lines through the origin as well as compositions of both operations, can be written as

\begin{equation}
\eqlabel{pin_trafo_vec}
    \mvec{a}' = \pm \mvec{V} \mvec{a} \reverse{\mvec{V}},\; \mvec{V} \in \cliff{3}, \; \| \mvec{V} \| = 1.
\end{equation}

The definition can be extended to higher grade blades, by transforming each of its vectors individually. The transformation of a general blade can then be written as 

\begin{equation}
\eqlabel{pin_trafo}
    \bigwedge_{i=1}^k \mvec{a}_i \rightarrow \bigwedge_{i=1}^k \left( \pm \mvec{V} \mvec{a}_i \reverse{\mvec{V}} \right) = (\pm 1)^k \mvec{V} \left( \bigwedge_{i=1}^k \mvec{a}_i \right) \reverse{\mvec{V}}.
\end{equation}

The property that the outer product of the transformed vectors is equal to the transformed outer product makes any orthogonal transformation a so called \emph{outermorphism}. As a consequence, orthogonal transformations preserve the grade of blades, which geometrically translates to the fact that vectors are transformed to vectors, bivectors to bivectors and so on, which is what we would expect from a geometric transformation (a vector will not become a volume when rotated). A formal proof of this property can be found in \cite{dorst2007geometric}.

The multivector $\mvec{V}$ in \eqref{pin_trafo_vec} is called a \emph{versor}. Versors together with the geometric product form a group on their own, the so called $\mathrm{Pin}(3)$ group. This group is said to be the \emph{double cover} of $O(3)$ (see \cite{lounesto2001clifford}). \eqref{pin_trafo} is both a representation of $O(3)$ and $\mathrm{Pin}(3)$. One can generalize this definition to arbitrary geometric algebras in the following way:

\begin{mathdef}{The \text{Pin(n,m,l)} and \text{Spin(n,m,l)} groups}

    Let $\mathcal{V} = \{v \in \cliff{n}{m}{l} \, | \, \|\mvec{v}\| = 1\}$ be the set of versors . Then $\mathcal{V}$ together with the geometric product forms the group $\mathrm{Pin}(n,m,l)$.

    Let $\mathcal{V} = \{v \in \cliff{n}{m}{l}^+ \, | \, \|\mvec{v}\| = 1\}$ be the set of even versors, i.e. versors that only involve even grades. Then $\mathcal{V}$ together with the geometric product forms the group $\mathrm{Spin}(n,m,l)$.
\end{mathdef}

\paragraph{The projective geometric algebra \texorpdfstring{$\cliff{3}{0}{1}$}{Cl(3,0,1)}}\mbox{}\\
The Euclidean geometric algebra $\cliff{3}$ allows for a powerful description of geometric objects in 3D space. However, it lacks the ability to describe absolute positions. For example, planes parametrized by 2-blades are restricted to go through the origin, and vectors only describe a relative displacement, not points in space. These problems are solved by the projective geometric algebra (PGA), which is the geometric algebra $\cliff{3}{0}{1}$ with three \emph{Euclidean} basis vectors squaring to $1$, $\mvec{e}_1^2 = \mvec{e}_2^2 = \mvec{e}_3^2 = 1$ and one \emph{null} vector squaring to $0$, $\mvec{e}_0^2 = 0$. We thus use an algebra based on a four dimensional vector space to describe 3D space, similarly to what one does with homogeneous coordinates. In the following we will denote PGA vectors which only contain a Euclidean part by $\evec{a}$.

There are multiple ways to interpret the different algebraic elements geometrically. Here, we will focus on the plane-based approach as described in \cite{dorst2020guided}, since this yields the most useful description of Euclidean motions as orthogonal transformations. Elements of grade $1$ are interpreted not as points but as planes, i.e. a vector of the form

\begin{equation}
    \eqlabel{pga_vector}
    \mvec{n} = \evec{n} + \delta \mvec{e}_0
\end{equation}

corresponds to a plane with (Euclidean) normal vector $\evec{n}$ and a distance $\delta$ from the origin. In turn, the Euclidean vectors $\mvec{e}_1, \mvec{e}_2, \mvec{e}_3$ correspond to basis planes through the origin, whereas $\mvec{e}_0$ is interpreted as the plane at infinity. Notably any multiple of \eqref{pga_vector} of the form $\alpha (\evec{n} + \delta \mvec{e}_0), \; \alpha \in \mathbb{R}$ represents the same plane.

\begin{figure}[t]
    \centering
    \includegraphics[width=0.9\textwidth]{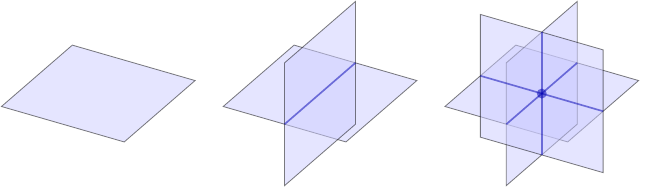}
    \caption{Visualization of the different geometric primitives in $\cliff{3}{0}{1}$. Vectors represent planes, 2-blades represent lines resulting from the intersecting of its generating planes and 3-blades represent points as the intersection of three planes.}
    \label{fig:pga_primitives}
\end{figure}

Similarly to the case of $\cliff{3}$, we can make use of \eqref{subspace_rep} to obtain a geometric interpretation for blades of higher grade. For 2-blades of the form $\mvec{L} = \mvec{a} \wedge \mvec{b}$, the subspace spanned by $\mvec{a}$ and $\mvec{b}$ corresponds to all planes which contain the line of intersection between the two original planes. It is thus sensible to take $\mvec{L}$ as representation of this line. Similarly, a 3-blade $\mvec{X} = \mvec{a} \wedge \mvec{b} \wedge \mvec{c}$ corresponds to the subspace spanned by all planes which contain the point of intersection of $\mvec{a}, \mvec{b}$, and $\mvec{c}$, and thus 3-blades represent points in PGA. The logic behind this construction is visualized in \figref{pga_primitives}. 

\begin{table}[t]
    \caption{Basis elements of the projective geometric algebra along with their geometric interpretation}
    \vspace{4pt}
    \centering
    \begin{tabular}{ccc}
        \toprule
        Grade & Basis Elements & Geometric Interpretation \\
        \midrule
        0 & $\{1\}$ & scalar \\ 
        1 & $\{\gavec{e_0}, \gavec{e_1}, \gavec{e_2}, \gavec{e_3}\}$ & planes \\
        2 & $\{\gavec{e_0}\gavec{e_1}, \gavec{e_0}\gavec{e_2}, \gavec{e_0}\gavec{e_3}, \gavec{e_1}\gavec{e_2}, \gavec{e_1}\gavec{e_3}, \gavec{e_2}\gavec{e_3}\}$ & lines, vanishing lines \\
        3 & $\{\gavec{e_0}\gavec{e_1}\gavec{e_2}, \gavec{e_0}\gavec{e_1}\gavec{e_3}, \gavec{e_0}\gavec{e_2}\gavec{e_3}, \gavec{e_1}\gavec{e_2}\gavec{e_3}\}$ & points, vanishing points \\
        4 & $\{\gavec{e_0}\gavec{e_1}\gavec{e_2}\gavec{e_3}\}$ & pseudoscalar \\
        \bottomrule
    \end{tabular}
    \label{tab:pga_elements}
\end{table}

PGA also contains additional classes of geometric objects which e.g. result from taking the outer product of parallel planes. For a further discussion of these, we refer to \cite{dorst2020guided}.

\subparagraph*{Meet and Join}

We have seen that the outer product in PGA can be used to calculate the intersection of geometric objects. In this context it is thus also known as the so called \emph{meet}. There is also an opposite operation called the \emph{join}\footnote{In some literature the roles of meet and join are actually reversed, i.e. the outer product takes the role of the join. This seeming contradiction can be explained by the fact that we follow the approach of interpreting vectors as planes and not as points.}, which, as the name suggests, e.g. maps two points to the line which passes through both of them. In order to properly define the join we first extend the concept of duality from $\cliff{3}$ to $\cliff{3}{0}{1}$.

\begin{mathdef}{Hodge duality}

    Let $\mvec{X} \in \cliff{3}{0}{1}$ be a blade. Then its \emph{hodge dual}  $\star\mvec{X}$ is defined via

    \begin{equation}
        \mvec{X} \, {\star\mvec{X}} = \left(\mvec{X}_e \reverse{\mvec{X}_e}\right) \pgaI,
    \end{equation}

    where $\mvec{X}_e$ is the Euclidean part of $\mvec{X}$ without $\mvec{e}_0$. For general multivectors the hodge dual is performed bladewise.
\end{mathdef}

As explained in \cite{dorst2020guided}, this definition is slightly different compared to \defref{ega_duality} since in PGA $\mvec{e}_0$ does not have a multiplicative inverse. However, the overall idea to map between the different subspaces of equal dimensionality is still the same. In practice the hodge dual maps a basis element simply to the element which contains all the basis vectors not present in the initial multivector (in some situations with an additional minus sign), e.g. $\star \mvec{e}_0\mvec{e}_3 = \mvec{e}_1\mvec{e}_2$ (confer \cite{dorst2020guided} for more details). Using the hodge dual one can define the \emph{join} as follows.

\begin{mathdef}{Join}
    \deflabel{join}

Let $\mvec{A}, \mvec{B} \in \cliff{3}{0}{1}$, then the join between these multivectors is defined as
    \begin{equation}
        \mvec{A} \vee \mvec{B} = \star^{-1}(\star\mvec{B} \wedge \star\mvec{A}),
    \end{equation}
where $\star^{-1}$ is the inverse hodge dual, which differs from the usual hodge dual by a sign for some elements.
\end{mathdef}

As mentioned earlier, the join can be viewed as being dual to the meet, linking geometric objects together instead of finding their incidence. In that way, the join of two points is the line connecting both of them and the join of three points results in a plane containing all three points.

In PGA a norm for multivectors can be defined analogously to \defref{ga_norm}. However, since PGA contains the null vector $\mvec{e}_0$, only half of the components of a general multivector contribute to the value of the norm. It is thus useful to define a second norm which depends on all of the components which contain $\mvec{e}_0$. 

\begin{mathdef}{Infinity norm}
\deflabel{infinity_norm}

Let $\mvec{A} \in \cliff{3}{0}{1}$, then its \emph{infinity norm} is given by

\begin{equation}
    \|\mvec{A}\|_{\infty} = \|{\star\mvec{A}}\|.
\end{equation}

\end{mathdef}

\subparagraph*{Euclidean transformations}

Similarly to the case of $\cliff{3}$, one can write the reflection of a plane $\mvec{m}$ in another plane $\mvec{p}$ as

\begin{equation}
    \mvec{m}' = - \mvec{p} \mvec{m} \mvec{p}, \; \|\mvec{p}\| = 1.
\end{equation}

Reflection of higher grade blades, e.g. lines and points, can again be achieved by reflecting each vector in the blade individually and making use of the fact that the reflection is an outermorphism

\begin{equation}
    \eqlabel{pga_reflection}
    \bigwedge_{i=1}^k \mvec{a}_i \rightarrow \bigwedge_{i=1}^k \left( -\mvec{p} \mvec{a}_i \mvec{p} \right) = (-1)^k \mvec{p} \left( \bigwedge_{i=1}^k \mvec{a}_i \right) \mvec{p}.
\end{equation}

The crucial difference in comparison to $\cliff{3}$ is that the reflecting plane is no longer restricted to go through the origin. This allows one to do consecutive reflections not only in intersecting planes, but also in parallel planes. Two important special cases are the reflections in two intersecting planes and the reflecions in two parallel planes as shwon in \figref{pga_trafos}. The first case corresponds to a rotation around the line of intersection by an angle which is twice the angle enclosed by the planes. The second case results in a translation along the distance vector by an amount equal to twice the distance between the planes (see \cite{dorst2020guided}).

\begin{figure}
    \centering
    \includegraphics[width=0.7\textwidth]{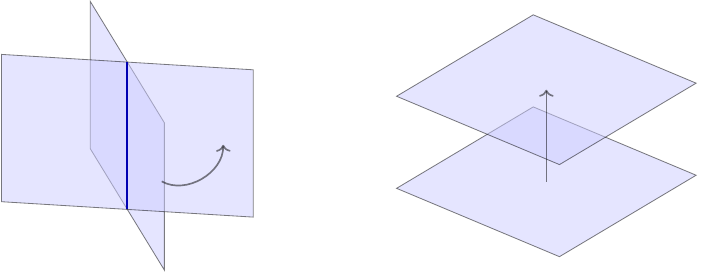}
    \caption{Visualization of the different geometric transformations in $\cliff{3}{0}{1}$}
    \figlabel{pga_trafos}
\end{figure}

Compositions of these two types of transformations make up the special Euclidean group $\mathrm{SE}(3)$, which can be seen as the group of rigid body motions. This means that every Euclidean transformation which does not involve reflections can be embedded as an element of the even subalgebra $\pga^+ = \{A \in \pga \, | \, \langle A \rangle_k = 0, \, k \, \text{odd}\}$, as stated in the following theorem:

\begin{theorem}{Euclidean motors}

    Let $T \in \mathrm{SE}(3)$ be an element of the special Euclidean group. Then there exists a multivector called \emph{motor} in the even subalgebra $\mvec{M} \in \cliff{3}{0}{1}^+$ with $\|\mvec{M}\| = 1$ such that 

    \begin{equation}
        \eqlabel{pga_motor}
        \mvec{X}' =  \mvec{M} \mvec{X} \reverse{\mvec{M}},
    \end{equation}

    with an arbitrary multivector $\mvec{X} \in \cliff{3}{0}{1}$, is a representation of $T$ on $\cliff{3}{0}{1}$.
\end{theorem}

We emphasize that the expression in \eqref{pga_motor} can be applied to any geometric object which is representable in the algebra, i.e. it does not matter if $\mvec{X}$ is a point, a vector, or a plane; the correct equation to transform it always takes the above form. This remarkable property can also be found for other important operations in PGA. In fact, we have already mentioned the \emph{meet} operation which allows to calculate intersections of arbitrary geometric objects $\mvec{A}, \mvec{B} \in \pga$ via the universal formula $\mvec{A} \wedge \mvec{B}$, as well as the \emph{join} which can likewise be applied to pairs of arbitrary objects.

\subparagraph*{Metric Relations}

We have already seen how to transform objects, find incidences between them, and join them together. Furthermore the basic operations of PGA allow to calculate a host of different metric relations between its elements. Below we provide a non-complete list of examples. 

\begin{theorem}{Metric relations}

Let $\mvec{p}_1, \mvec{p}_2 \in \pga^{[1]}$ be planes, $\mvec{L} \in \pga^{[2]}$ be a line and  $\mvec{P}_1, \mvec{P}_2 \in \pga^{[3]}$be points, which are all normalized, i.e. 
\begin{equation*}
    \|\mvec{p}_1\| = \|\mvec{p}_2\| = \|\mvec{L}\|= \|\mvec{P}_1\| = \|\mvec{P}_2\| = 1,
\end{equation*}
then we can calculate the following relations:

\begin{itemize}
    \item Distance between points $\mvec{P}_1, \mvec{P}_2$

    \begin{equation}
        \|\mvec{P}_1 \vee \mvec{P}_2\|
    \end{equation}

    \item Distance between point $\mvec{P}_1$ and line $\mvec{L}$

    \begin{equation}
        \|\mvec{P}_1 \vee \mvec{L}\|
    \end{equation}

    \item Distance between point $\mvec{P}_1$ and plane $\mvec{p}_1$

    \begin{equation}
        \|\mvec{P}_1 \wedge \mvec{p}_1\|_\infty
    \end{equation}

    \item Angle between planes $\mvec{p}_1, \mvec{p}_2$

    \begin{equation}
        \sin^{-1}\left(\|\mvec{p}_1 \wedge \mvec{p}_2\|\right)
    \end{equation}

    \item Angle between plane $\mvec{p}_1$ and line $\mvec{L}$

    \begin{equation}
        \sin^{-1}\left(\|\grade{\mvec{p}_1\mvec{L}}{3}\|\right)
    \end{equation}
\end{itemize}    
\end{theorem}

For an extensive list of operations see \cite{dorst2020guided}.

\subsection{Methodology}
\subsubsection{GAFL architecture}
\seclabel{framediff_architecture}

\begin{figure}[h]
    \centering
    \includegraphics{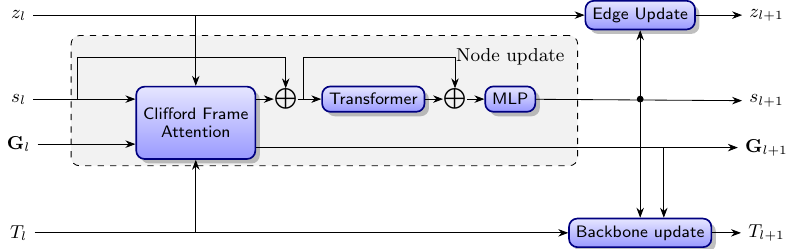}
    \caption{High level overview over the GAFL architecture. The architecture was adapted from FrameDiff/FrameFlow \cite{yim2023framediff, yim2023frameflow}. We replaced invariant point attention (IPA) with Clifford frame attention (CFA) and added geometric node features $\mvec{G}$.}
    \label{fig:gafl_flowchart}
\end{figure}

We adapted the GAFL architecture from FrameDiff/FrameFlow \cite{yim2023framediff, yim2023frameflow} and replaced invariant point attention (IPA) with Clifford frame attention (CFA) (see Algorithm \ref{alg:GA_IPA}). Furthermore we introduced geometric node features $\mvec{G}_i$ that are used in the prediction of geometric attention values and backbone frame updates. An overview over the architecture is shown in Figure~\ref{fig:gafl_flowchart}. The backbone update block is presented in Algorithm \ref{alg:backbone_update_modified}. For details on the edge update we refer to \cite{yim2023framediff}.

\subsubsection{Invariant Point Attention}
\label{sec:originalipa}

In the following we provide a short overview over the IPA architecture.

\begin{figure}
    \centering
    \includegraphics[width=0.7\linewidth]{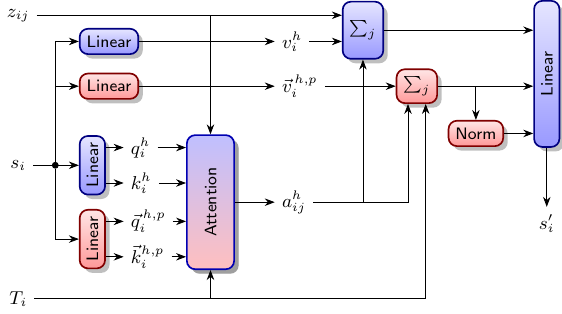}
    \caption{High level overview of invariant point attention \cite{jumper2021highly}. Blue nodes represent layers which process scalar information, while red nodes represent layers which process geometric information.}
    \label{fig:ipa}
\end{figure}

\begin{algorithm}[h]
    \caption{Invariant point attention (IPA)~\citep{jumper2021highly}}
\begin{algorithmic}[1]
    \Procedure{IPA}{$\left\{\mathbf{s}_i\right\},\left\{\mathbf{z}_{i j}\right\},\left\{T_i\right\}, N_{\text {head }} = 8, c=128, N_{\text {query points }}=8, N_{\text {point values }}=12$}
    \State $\mathbf{q}_i^h, \mathbf{k}_i^h, \mathbf{v}_i^h=\operatorname{LinearNoBias}\left(\mathbf{s}_i\right)$ \Comment{$\mathbf{q}_i^h, \mathbf{k}_i^h, \mathbf{v}_i^h \in \mathbb{R}^c, h \in\left\{1, \ldots, N_{\text {head }}\right\}$}
\State $\vec{\mathbf{q}}_i^{h p}, \vec{\mathbf{k}}_i^{h p}=\text { LinearNoBias }\left(\mathbf{s}_i\right)$ \Comment{$\vec{\mathbf{q}}_i^{h p}, \vec{\mathbf{k}}_i^{h p}, \in \mathbb{R}^3, p \in\left\{1, \ldots, N_{\text {query points }}\right\}$}
\State $\vec{\mathbf{v}}_i^{h p}=\operatorname{LinearNoBias}\left(\mathbf{s}_i\right)$ \Comment{$\vec{\mathbf{v}}_i^{h p} \in \mathbb{R}^3, p \in\left\{1, \ldots, N_{\text {point values }}\right\}$}
\State $b_{i j}^h=\operatorname{LinearNoBias}\left(\mathbf{z}_{i j}\right)$
\State $w_C=\sqrt{\frac{2}{9 N_{\text {query points }}}} \text {, }$
\State $w_L=\sqrt{\frac{1}{3}}$
\State $a_{i j}^h=\operatorname{softmax}_j\left(w_L\left(\frac{1}{\sqrt{c}} \mathbf{q}_i^{h^{\top}} \mathbf{k}_j^h+b_{i j}^h-\frac{\gamma^h w_C}{2} \sum_p\left\|T_i \circ \vec{\mathbf{q}}_i^{h p}-T_j \circ \vec{\mathbf{k}}_j^{h p}\right\|^2\right)\right)$
\State $\tilde{\mathbf{o}}_i^h=\sum_j a_{i j}^h \mathbf{z}_{i j}$
\State $\mathbf{o}_i^h=\sum_j a_{i j}^h \mathbf{v}_j^h$
\State $\vec{\mathbf{o}}_i^{h p}=T_i^{-1} \circ \sum_j a_{i j}^h\left(T_j \circ \vec{\mathbf{v}}_j^{h p}\right)$
\State $\tilde{\mathbf{s}}_i=\operatorname{Linear}\left(\operatorname{concat}_{h, p}\left(\tilde{\mathbf{o}}_i^h, \mathbf{o}_i^h, \vec{\mathbf{o}}_i^{h p},\left\|\vec{\mathbf{o}}_i^{h p}\right\|\right)\right)$
\State \textbf{return} $\left\{\tilde{\mathbf{s}}_i\right\}$
    \EndProcedure
    \end{algorithmic}
    \label{alg:ipa}
\end{algorithm}

\begin{algorithm}[h]
    \caption{Original backbone update~\citep{jumper2021highly}}
\begin{algorithmic}[1]
    \Procedure{BackboneUpdate}{$\{\mathbf{s}_i\}, \{T_i\}$}
    \State $b_i, c_i, d_i, \vec{\mathbf{t}_i} = \operatorname{Linear}(\mathbf{s}_i)$
    \State $(a_i, b_i, c_i, d_i) = (1, b_i, c_i, d_i) / \sqrt{1 + b_i^2 + c_i^2 + d_i^2}$
    \State $R_i = \operatorname{QuaternionToMatrix}(a_i, b_i, c_i, d_i)$
    \State $\tilde{T}_i = (R_i, \vec{\mathbf{t}_i})$
    \State \textbf{return} $T_i \circ \tilde{T}_i$
    \EndProcedure
    \end{algorithmic}
\label{alg:backbone_update}
\end{algorithm}

Invariant point attention uses local frames to construct the messages between nodes, where the coordinate frames are given by the frames of the individual residues. The full procedure is presented in algorithm \ref{alg:ipa} and visualized in \figref{ipa}. The part of IPA that processes geometric information (red nodes in \figref{ipa}) can be summarized as follows:

\begin{enumerate}
    \item Learn a certain number of \emph{local} vector valued queries, keys and values $\vec{\mathbf{q}}_i, \vec{\mathbf{k}}_i, \vec{\mathbf{v}}_i$.
    \item In the calculation of attention scores, transform the local queries and keys into the global frame via $T_i \circ \vec{\mathbf{q}}_i$ and calculate the squared distance between them. Importantly they should be thought of as points in 3D space rather than vectors, meaning that they change under translations, which vectors would not, and also making the $L^2$-norm of their separation a somewhat natural map to construct invariant attention scores.
    \item In the message passing step, transform vector valued features from the neighboring frame $j$ to the node frame $i$ via $T_i^{-1} \circ T_j \circ \vec{\mathbf{q}}_i$ and aggregate the vector valued information over all neighbors.
    \item Finally, concatenate the output vectors along with their norm with the remaining scalar output features and put them through a final linear layer.
\end{enumerate}

The update of the backbone frames, which is used in conjunction with IPA e.g. in \cite{jumper2021highly, yim2023frameflow}, is accomplished by learning an $\text{SE}(3)$ transformation per frame which is concatenated with the current frame representation. To this end, the network predicts a quaternion for the rotation and a translation vector from the scalar node features $\mathbf{s}_i$ as shown in algorithm \ref{alg:backbone_update}

\subsubsection{Clifford frame attention}
\label{sec:cfa-method}

\begin{figure}[h]
    \centering
    \includegraphics{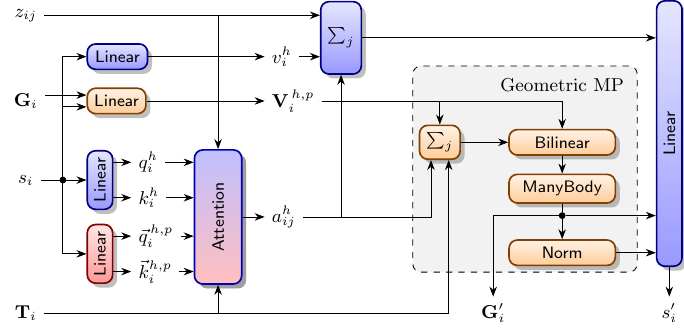}
    \caption{Overview of Clifford frame attention. Blue nodes represent layers which process scalar information, red nodes represent layers which process point valued information and orange nodes represent layers which process features in the PGA. The central innovation is the novel construction of geometric messages, summarized in the grey box. To retain readability we omit the calculation of relative frame transformations in the chart. }
    \label{fig:ipa_ga}
\end{figure}

\begin{algorithm}[h]
    \caption{Clifford frame attention}
    \alglabel{GA_IPA}
\begin{algorithmic}[1]
    \Procedure{CFA}{$\left\{\multivector{s}_i\right\}, \left\{\multivector{G}_i\right\},\left\{\multivector{z}_{i j}\right\},\left\{T_i\right\}, N_{\text {head }} = 8, c=128, N_{\text {query points }}=8, N_{\text {point values }}=4$}
    \State $\multivector{q}_i^h, \multivector{k}_i^h, \multivector{v}_i^h=\operatorname{LinearNoBias}\left(\multivector{s}_i\right)$ \Comment{$\multivector{q}_i^h, \multivector{k}_i^h, \multivector{v}_i^h \in \mathbb{R}^c, h \in\left\{1, \ldots, N_{\text {head }}\right\}$}
\State $\myvector{{q}}_i^{h p}, \myvector{{k}}_i^{h p}=\text { LinearNoBias }\left(\multivector{s}_i\right)$ \Comment{$\myvector{{q}}_i^{h p}, \myvector{{k}}_i^{h p}, \in \mathbb{R}^3, p \in\left\{1, \ldots, N_{\text {query points }}\right\}$}
\State $\color{dblue}\multivector{V}_i^{p}=\operatorname{Linear}\left(\multivector{s}_i\right)$ \Comment{$\color{dblue}\multivector{V}_i^{p} \in \mathbb{R}^{16}, p \in\left\{1, \ldots, N_{\text {point values }}\right\}$}
\State $\color{dblue}\multivector{V}_i^{h p}=\operatorname{EquiLinear}\left(\operatorname{concat}_p\left(\multivector{V}_i^{p}, \multivector{G}_i^{p}\right)\right)$
\State $b_{i j}^h=\operatorname{LinearNoBias}\left(\multivector{z}_{i j}\right)$
\State $w_C=\sqrt{\frac{2}{9 N_{\text {query points }}}} \text {, }$
\State $w_L=\sqrt{\frac{1}{3}}$
\State $a_{i j}^h=\operatorname{softmax}_j\left(w_L\left(\frac{1}{\sqrt{c}} \multivector{q}_i^{h^{\top}} \multivector{k}_j^h+b_{i j}^h-\frac{\gamma^h w_C}{2} \sum_p\left\|T_i \circ \myvector{{q}}_i^{h p}-T_j \circ \myvector{{k}}_j^{h p}\right\|^2\right)\right)$
\State $\tilde{\multivector{o}}_i^h=\sum_j a_{i j}^h \multivector{z}_{i j}$
\State $\multivector{o}_i^h=\sum_j a_{i j}^h \multivector{v}_j^h$
\State $\color{dblue} \multivector{T}^{\text{rel}\, h}_i = \sum_j a_{ij}^h \multivector{T}_i^{-1} \multivector{T}_j$
\State $\color{dblue}\tilde{\multivector{V}}_i^{h p}=\operatorname{EquiLinear}\left(\operatorname{concat}\left(\multivector{V}_i^{h p}, \multivector{T}^{\text{rel}\, h}_i\right)\right)$
\State $\color{dblue}\multivector{O}_i^{h p}=\operatorname{GeometricBilinear\Big(\multivector{T}_i^{-1}\sum_j a_{ij} \left(\multivector{T}_j\tilde{\multivector{V}}^{hp}_j\multivector{T}_j^{-1}\Big)\multivector{T}_i, \, \multivector{V}^{hp}_i\right)}$
\State $\color{dblue}\multivector{O}_i^{h p}=\operatorname{GeometricManyBodyProduct}(\multivector{O}_i^{h p})$
\State $\color{dblue}\tilde{\multivector{s}}_i=\operatorname{Linear}\left(\operatorname{concat}_{h, p}\left(\tilde{\multivector{o}}_i^h, \multivector{o}_i^h, \multivector{O}_i^{h p},\left\|\multivector{O}_i^{h p}\right\|, \left\|\multivector{O}_i^{h p}\right\|_{\infty},\multivector{T}^{\text{rel}\, h}_i\right)\right)$
\State $\color{dblue}\tilde{\multivector{G}}_i=\operatorname{EquiLinear}(\multivector{O}_i^{h p})$
\State \textbf{return} $\color{dblue}\left\{\tilde{\multivector{s}}_i, \tilde{\multivector{G}}_i,\multivector{T}^\text{rel}_i\right\}$
    \EndProcedure
    \end{algorithmic}
\label{alg:ipa_modified}
\end{algorithm}

\begin{algorithm}
    \caption{GeometricBilinear \cite{brehmer2023geometric}}
\begin{algorithmic}[1]
    \Procedure{GeometricBilinear}{$\multivector{A}, \multivector{B}$}
    \State $\multivector{G}_L = \operatorname{EquiLinear}\left(\multivector{A}\right)$
    \State $\multivector{G}_R = \operatorname{EquiLinear}\left(\multivector{B}\right)$
    \State $\multivector{G} = \operatorname{GeometricProduct}(\multivector{G}_L, \multivector{G}_R)$ 
    \State $\multivector{J}_L = \operatorname{EquiLinear}\left(\multivector{A}\right)$
    \State $\multivector{J}_R = \operatorname{EquiLinear}\left(\multivector{B}\right)$
    \State $\multivector{J} = \operatorname{Join}(\multivector{J}_L, \multivector{J}_R)$
    \State \textbf{return} $\operatorname{EquiLinear}\left(\operatorname{concat}\left(\multivector{G}, \multivector{J}\right)\right)$
    \EndProcedure
    \end{algorithmic}
    \label{alg:pairwisegeometricbilinear}
\end{algorithm}

\begin{algorithm}[h]
    \caption{GeometricManyBodyProduct}
\begin{algorithmic}[1]
    \Procedure{GeometricManyBodyProduct}{$\multivector{A}$, $W_{nijk}$, $\tilde{W}_{nijk}$} 
        \State $\multivector{X} = \operatorname{EquiLinear}(\multivector{A})$ \Comment{$\multivector{A}$ contains aggregated messages}
        \State $\multivector{Y} = \operatorname{EquiLinear}(\multivector{A})$
        \State $\mvec{O}_n = \left[\sum_{ijk} \left( W_{nijk} \bigg\langle\langle\mvec{X}_n\rangle_j \langle\mvec{Y}_n\rangle_k\bigg\rangle_i + \tilde{W}_{nijk} \bigg\langle\langle\mvec{X}_n\rangle_j \vee \langle\mvec{Y}_n\rangle_k\bigg\rangle_i\right)\right] + \mvec{Y}_n$
        \State \textbf{return} $\multivector{O}$
    \EndProcedure
    \\
    \Comment{$n$ is a feature dimension, $W$, $\tilde{W}$ are learnable, $\langle\multivector{X}\rangle_i$ is the projection onto the $i$-th grade of $\multivector{X}$}
    \end{algorithmic}
    \label{alg:geometricmanybodycontraction}
\end{algorithm}

\begin{algorithm}
    \caption{Backbone update}
\begin{algorithmic}[1]
    \Procedure{BackboneUpdate}{$\{\multivector{s}_i\}, \{\multivector{G}_i\}, \{\multivector{T}^\text{rel}_i\}, \{\multivector{T}_i\}$}
    \State $\color{dblue} b_i, c_i, d_i, \vec{\mathbf{t}_i} = \operatorname{MLP}(\operatorname{concat}\left(\multivector{s}_i, \multivector{G}_i, \multivector{T}^\text{rel}_i)\right)$
    \State $\color{dblue} \multivector{R}_i = \operatorname{EmbedRotor(b_i, c_i, d_i)}$
    \State $\color{dblue} \multivector{S}_i = \operatorname{EmbedTranslator(\vec{\mathbf{t}_i})}$
    \State \textbf{return} $\color{dblue} \multivector{T}_i  \multivector{R}_i \multivector{S}_i$
    \EndProcedure
    \end{algorithmic}
\label{alg:backbone_update_modified}
\end{algorithm}

An overview over the CFA architecture is given in Figure \ref{fig:ipa_ga}. The proposed changes to the original IPA and FrameDiff architecture are shown in algorithms \ref{alg:ipa_modified} and \ref{alg:backbone_update_modified}, highlighted in blue. They can be summarized as follows:

\begin{enumerate}
    \item We replace point-valued attention values with multivectors $\multivector{V}_i^{h p}$ and also introduce geometric node features $\mvec{G}_i \in \pga$.
    \item We compute node features $\multivector{T}^\text{rel}$ containing aggregated relative transformations between frames $\multivector{T}_i$ and $\multivector{T}_j$ as
    
    \begin{equation}
        \multivector{T}^\text{rel}_i \equiv \sum_{j}a_{ij} \multivector{T}^{-1}_i\multivector{T}_j
        \label{eq:T-rel}
    \end{equation}
        
    These features are concatenated with the remaining geometric features and also passed directly to the backbone update block via a residual connection.
    
    \item Messages are formed by applying geometric bilinear layers from \citep{brehmer2023geometric} to node features of each node pair, as described in algorithm \ref{alg:pairwisegeometricbilinear}. The $\operatorname{EquiLinear}$ layer refers to the most general equivariant linear layer in PGA, also introduced in \citep{brehmer2023geometric}, which for input features $\mvec{X}_n$ can be written as
    \begin{equation}
    \eqlabel{pga_linear}
    \operatorname{EquiLinear}(\mvec{X}_n) = \sum_m \left[ \sum_{k=0}^{4}w_{knm} \langle \mvec{X}_{m} \rangle_k + \sum_{k=0}^3 v_{knm} \mvec{e}_0 \langle \mvec{X}_{m} \rangle_k \right].
    \end{equation}
    Although we do not need this property for enforcing equivariance, we use it for its parameter efficiency and geometric inductive bias.
    \item We construct implicit higher body messages inspired by MACE~\cite{batatia2022mace}.
    Bilinearity of the geometric product and the join means that repeated products of aggregated two-body messages correspond to a sum of $N$-body messages as shown in eq.~\ref{eq:higher_order}. In algorithm~\ref{alg:geometricmanybodycontraction}, we use the geometric product and the join to construct three-body messages from aggregated two-body messages, stored in node features $\multivector{A}$.
    
    \item In addition to the Euclidean norm we also compute the infinity norm (see Definition \ref{def:infinity_norm}) of multivector features after message passing.

    \item In the Backbone update step (Algorithm \ref{alg:backbone_update_modified}), we concatenate scalar and geometric features along the feature dimension and pass them through an MLP to predict multivectors $\multivector{R}$ and $\multivector{S}$ that parameterize a rotation and translation respectively. These are then multiplied with the current frames to compute frame updates. The main difference to the backbone update as used in \cite{jumper2021highly, yim2023frameflow} is the inclusion of geometric features and the use of the MLP. The prediction of rotor and translator instead of a rotation matrix and translation vector is just a matter of representation.
\end{enumerate}

\subsubsection{Equivariance of GAFL}
\label{sec:gafl_equivariance}

As in the original IPA formulation, SE(3) equivariance of the GAFL architecture is based on expressing geometric features in canonically induced local frames, as described in section \secref{methods}.
    
More specifically, going through the architecture step by step, the attention scores are calculated using the $L_2$ norm of the difference of point features, which is E(3) invariant.
Equivariance of the message aggregation step is guaranteed by the expression  $$T_i^{-1} \circ T_j \circ \vec{v}_j^{hp}$$ in line 11 of algorithm 1, where $\vec{v}_j^{hp}$ are SE(3) invariant point values and the frames $\{T_i\}$ transform according to $T_i \rightarrow T_{global} \circ T_i$ such that the whole expression remains invariant: $$T_i^{-1} \circ T_{global}^{-1} \circ T_{global}\circ T_j \circ \vec{v}_j^{hp} = T_i^{-1} \circ T_j \circ \vec{v}_j^{hp}.$$ In GAFL, we do not modify the calculation of attention scores from IPA, which means that their invariance remains ensured.
During message passing, we use the same construction as above (see line 11 of Algorithm \ref{alg:ipa_modified}), but use a different representation of SE(3), namely multivector features instead of point features.
The choice of representation, however, does not influence the invariance of the whole expression.
Also the relative frame transformations $\mathbf{T}_i^{-1} \mathbf{T}_j$  we compute are invariant. All subsequent layers including the GeometricBilinear layer and the ManyBodyProduct layer operate exclusively on invariant node features, hence overall equivariance is retained throughout those layers.
Finally, in the backbone update step, we predict an invariant frame update, just like in IPA, which when concatenated with the original frame transforms equivariantly: $$\mathbf{T}_i\mathbf{T}_{update} \rightarrow \mathbf{T}_{global}\mathbf{T}_i\mathbf{T}_{update}.$$ 

Permutation equivariance is also maintained by the GAFL architecture, since we use message passing on the fully connected graph.
In the setting at hand, however, we break this permutation equivariance intentionally by introducing positional encodings for the nodes, as done in models that rely on original IPA such as RFdiffusion.

\subsection{Experiments}

\subsubsection{Definition of novelty and diversity}
\label{sec:novely_definition}

For each sampled length, we compute the pairwise Template Modeling ($\text{TM}$) score between designable backbones as a measure for similarity between folds~\citep{zhang2004scoring} and report \emph{diversity} as the $\text{TM}$ score averaged over all lengths,

\[\text{\textit{Diversity}} = \frac{1}{N} \sum_{l=1}^{N} \frac{1}{n_l(n_l-1)} \sum_{i=1}^{n_l} \sum_{\substack{j=1 \\ j \neq i}}^{n_l} \text{TM}(\mathbf{{x}}_i, \mathbf{{x}}_j)\] with the total number \( N \)  of sampled lengths, the number of designable samples $n_l$ at a given length, and the \( \mathbf{x}_i \) and \( \mathbf{x}_j \) being the \( i \)-th and \( j \)-th samples, respectively.

The resulting \textit{diversity} score reports on how similar are sampled backbones to each other. We compare the structures of designable samples to the natural proteins found in the PDB database using FoldSeek~\citep{van2024fast} in TMalign mode, and define \emph{novelty} as the average of the highest $\text{TM}$ score calculated over all designable samples:

\[
\text{\textit{Novelty}} = \frac{1}{n} \sum_{i=1}^{n} \max_{j} \, \text{TM}(\mathbf{x}_i, \mathbf{x}_j)
\] with 
\( n \) the number of all designable samples and \( \mathbf{x}_{j} \) being the sample from the PDB database with the highest \text{TM} score to the sample \( \mathbf{x}_{i} \)

Since diversity and novelty are defined as averaged similarity scores, small values are desirable for the generation of structurally distinct and \textit{de novo} proteins.

\subsubsection{Baselines used for comparison with GAFL}

\label{sec:baselines}

We generate backbones using RFdiffusion relying on publicly available weights with default settings (\textit{noise\_scale\_ca}:\,1)
(\href{https://github.com/RosettaCommons/RFdiffusion}{GitHub RFdiff}). For Genie, we use the model with the published weights trained either on the SCOPe or SwissProt datasets containing proteins of up to 128 or 256 residues long(\href{https://github.com/aqlaboratory/genie}{GitHub Genie}), and generate backbones with sampling for 1000 time steps, as it has demonstrated the best performance. FrameDiff was used with the newest published model weights (best$\_$weights.pth) with sampling for 500 timesteps (\href{https://github.com/jasonkyuyim/se3_diffusion}{GitHub FrameDiff}). FrameFlow was used with the published weights trained on the same dataset as GAFL (\href{https://github.com/microsoft/protein-frame-flow}{GitHub FrameFlow}). The originally model, denoted as FrameFlow$^*$ can be found at (\href{https://github.com/microsoft/protein-frame-flow/tree/legacy}{GitHub FrameFlow (legacy)}). We note that the GitHub repository supporting FrameFlow contains an implementation of minibatch OT~\cite{bose2023se}, which we use for retraining FrameFlow. For FoldFlow we use the optimal transport model (foldflow-ot.pth) with inference annealing as suggested by \cite{bose2023se} and sampling for 500 timesteps (\href{https://github.com/DreamFold/FoldFlow}{GitHub FoldFlow}).

\subsubsection{Results for small proteins}
In order to evaluate GAFL's performance for small proteins in particular, we evaluate a range of models in the setting from the original FrameFlow~\cite{yim2023frameflow} paper, where 10 backbones per length in $\{60,61,\ldots,128\}$ are generated.
Apart form the models in Table~\ref{tab:pdb-performance}, we also evaluate Genie~\cite{lin2023genie}, which was trained on backbones smaller than 300 residues and were therefore excluded from the evaluation in Table~\ref{tab:pdb-performance}.

We find that GAFL and FrameFlow are the only models capable of generating highly designable backbones with diverse secondary structures for the considered length range, where GAFL outperforms FrameFlow in designability and secondary structure content.
While FoldFlow and RFdiffusion, like GAFL, achieve designabilities over 90\%, they generate backbones with 0\% and 7\% $\beta$-strand content, respectively, indicating a mode-collapse towards generating $\alpha$-helical structures.
GAFL also outperforms both RFdiffusion and FoldFlow in terms of diversity and novelty.
All other models considered have significantly lower designability.
For Genie in particular, we observe good novelty but it has the lowest designability among the models considered and also under-represents beta strands.

\begin{table}
\centering
\caption{
Performance of different models when generating 10 backbones for each length in $\{60,61,\ldots,128\}$. For strand and helix content, we choose the PDB dataset filtered for lengths 60 to 128 as reference.
FrameFlow denotes the model trained on the PDB published in~\cite{yim2024improved}.
}
\label{tab:method_comparison1}
\vspace{0.1cm}
\resizebox{\textwidth}{!}{%
    \begin{tabular}{@{}lccccc@{}}
    \toprule
    Method           & Designability ($\uparrow$)  & Diversity ($\downarrow$) & Novelty ($\downarrow$) & Helix Content & Strand Content \\
    \midrule
    PDB (128)     & -         & -                 & -         & 0.36 (0.00)      & 0.22 (0.00)        \\ \midrule
    SCOPe(128)     & -         & -                 & -         & 0.33 (0.00)      & 0.26 (0.00)        \\ \midrule
    Genie-SCOPe      & 0.72 (0.02)         & 0.38 (0.00)              & \textbf{0.67} (0.00) & 0.66 (0.01)     & 0.07 (0.01)        \\
    Genie-SwissProt         & 0.79 (0.02)        & \textbf{0.37} (0.00)              & 0.68 (0.00)      & 0.66 (0.01)     & 0.09 (0.01)        \\
    FoldFlow-OT           & \textbf{0.99} (0.00)         & 0.49 (0.01)            & 0.83 (0.01)     & 0.87 (0.00)     & 0.00 (0.00)         \\
    FrameDiff               & 0.81 (0.02)          & 0.44 (0.00) & 0.72 (0.00)     & 0.39 (0.01)     & 0.30 (0.01)           \\
    FrameFlow  & 0.95 (0.01)          & 0.38 (0.00)             & 0.75 (0.00)     & 0.53 (0.01)      &  0.22 (0.01)             \\
    RFdiffusion             & 0.98 (0.01) & 0.43 (0.01)          & 0.8 (0.00)      & 0.78 (0.00)      & 0.07 (0.00)           \\
    \midrule
    GAFL            &0.96 (0.01)  & 0.38 (0.00)              & 0.78 (0.00)    & 0.49 (0.01)       & 0.27 (0.01)    \\
    \bottomrule
    \end{tabular}%
}
\end{table}

\subsubsection{Results for long proteins}
\label{sec:pdb_results}

We also evaluate the performance of GAFL for generating proteins up to length 500, which enables to compare GAFL against vector field networks (VFN) \cite{mao2023modeling}, which at the time of writing have no published model weights.
We thus evaluate GAFL and other baselines trained on the PDB with the same inference settings as in \cite{mao2023modeling}, sampling 5 protein backbones for each length in $\{100, 105, \dots, 500\}$.
We report the results in Table \ref{tab:GAFL_VFN_long}.
For VFN, we can only report on designability since \cite{mao2023modeling} have different definitions for diversity and novelty respectively and do not calculate the secondary structure content of generated structures.
\begin{table}[h]
\centering
\caption{
Comparison of GAFL with VFN, FrameDiff, and RFdiffusion for the generation of 5 protein backbones at each length $\in$ $
\{100, 105, \dots, 500\}$. Values for VFN from~\cite{mao2023modeling}.
}
\resizebox{\textwidth}{!}{%
    \begin{tabular}{@{}lcccccc@{}}
    \toprule
    Method            & Designability ($\uparrow$) & Diversity ($\downarrow$) & Novelty ($\downarrow$) & Helix Content & Strand Content \\
    \midrule
    FrameDiff         & 0.28 (0.02)               & 0.44 (0.01)                       & 0.70 (0.01) 
    & 0.57  & 0.17 \\
    VFN               & 0.44                & -                        & - & - & -                   \\
    RFdiffusion       & 0.71 (0.01)         &  0.36 (0.00)     & \textbf{0.68} (0.00)  & \textbf{0.52} & \textbf{0.28}             \\
    FrameFlow$^{*}$   & 0.64 (0.02)         &  \textbf{0.33} (0.00)     & 0.70 (0.01)  & 0.64 & 0.15             \\
    GAFL (Ours)       & \textbf{0.74} (0.01)      & 0.35 (0.00)            & 0.74 (0.00)  & 0.68 & 0.14           \\
    \bottomrule
    {\footnotesize{*Published model weights}}
    \end{tabular}%
}
\label{tab:GAFL_VFN_long}
\end{table}

We find that GAFL generates the highest fraction of designable backbones, while also yielding the highest diversity.
At the same time RFdiffusion has the best novelty and generates structures with a secondary structure content which is closest to the PDB.
The trend that GAFL outperforms RFdiffusion with respect to secondary structure content on small proteins but performs worse on longer proteins is also already apparent from Figure \ref{fig:des_by_len_combined}. 

\subsubsection{Ablation for CFA on the PDB}
\label{sec:pdb_ablation}

As described in Section~\ref{sec:ablations}, we retrained FrameFlow in the same setup as GAFL with default hyperparameters from the repository in~\ref{sec:baselines}.
For both FrameFlow and GAFL, we performed three training runs with the hyperparameters reported in Appendix \ref{sec:ckpt_hyperparams}. For all runs we selected checkpoints using our checkpointing criterion, described in Section \ref{sec:training_details}. In Table~\ref{tab:gafl_ff_PDB}, we report the performance of the best checkpoints per run.

\begin{figure}[h]
    \centering
    \includegraphics[page=1,width=.6\textwidth]{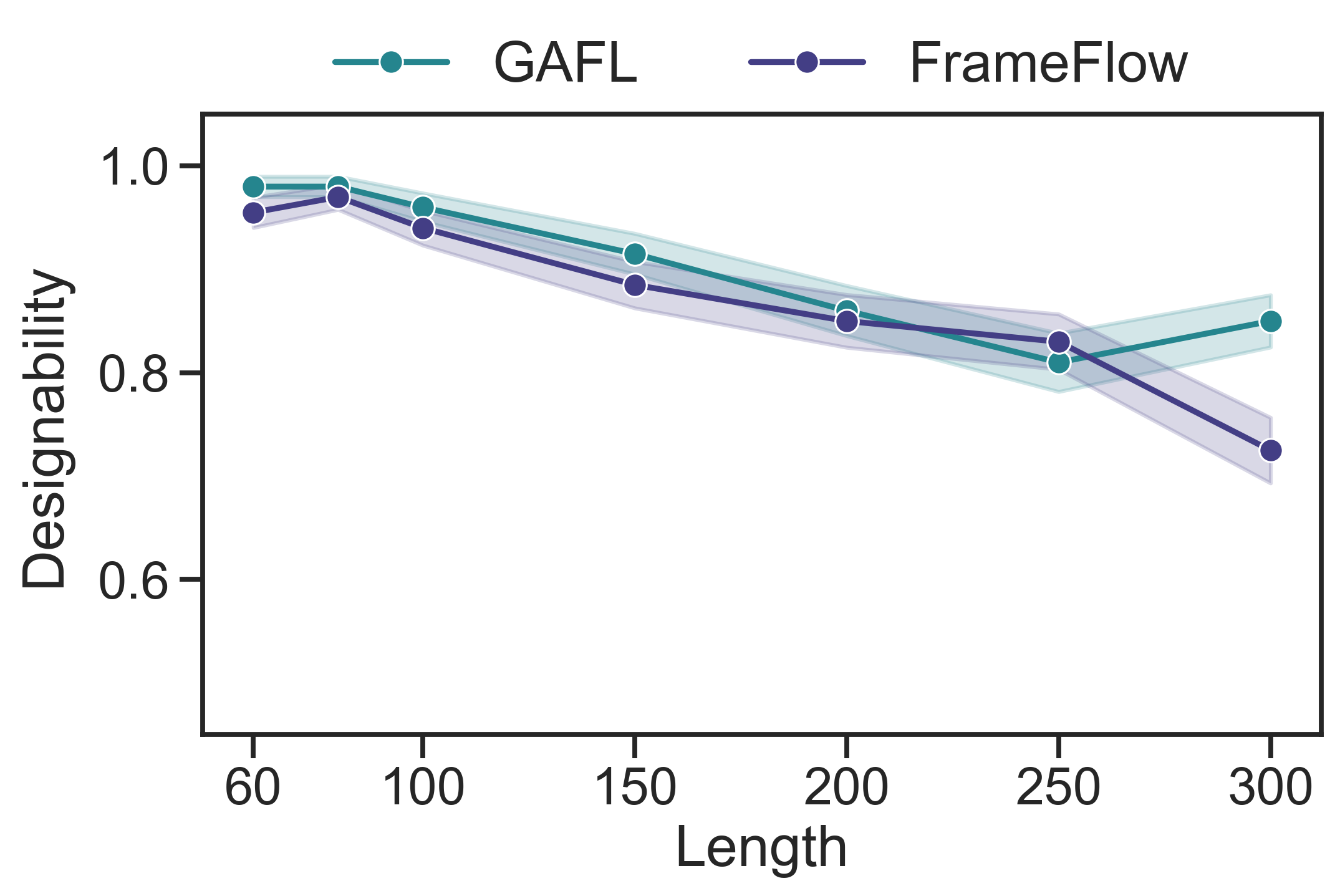}
    \caption{Designability of GAFL and the published FrameFlow model as a function of length with standard errors obtained by bootstrapping the set of generated samples. 200 backbones are generated for each length, as in Figure~\ref{fig:des_by_len_combined}.}
    \label{fig:gafl_ff_des_by_len}
\end{figure}

As extension of Figure~\ref{fig:des_by_len_combined}, we directly compare the designability of GAFL and the published FrameFlow model by length in Figure~\ref{fig:gafl_ff_des_by_len}.
We find that GAFL has higher designability for lengths smaller than 200 and greater than 250.
Since we only generate 200 backbones for each length, the margins of error of the individual designabilities overlap, except for length 300.
The total designability of GAFL, averaged over all 1000 backbones, however, is significantly better than that of FrameFlow (Table~\ref{tab:pdb-performance}).

\subsubsection{Timestep analysis of GAFL}

We perform a timestep analysis, evaluating the best GAFL checkpoint with a different number of timesteps taken during inference, in order to judge its impact on the designability metric. To this end we generate 100 backbones for each length in $\{100,150,\ldots,300\}$, for different numbers of timesteps. We see that the performance beyond 200 timesteps stays almost constant, suggesting that 200 timesteps represents a good tradeoff between performance and inference speed.

\begin{figure}[h]
    \centering
    \includegraphics[page=1,width=.6\textwidth]{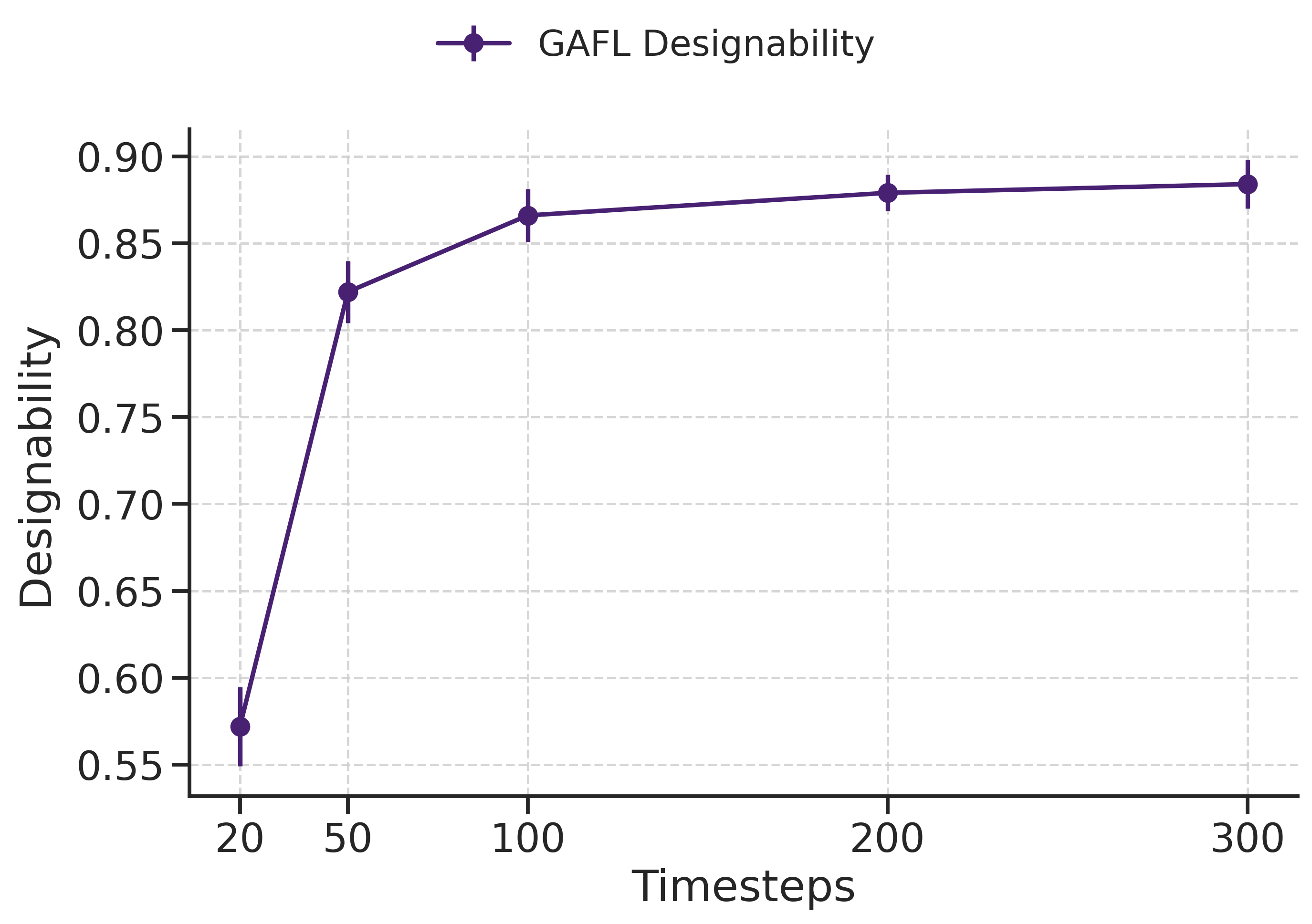}
    \caption{Designability as a function of timesteps used during inference. 100 backbones for each length in $\{100,150,\ldots,300\}$ were sampled per each data point. Vertical lines denote the standard deviation.}
    \label{fig:timesteps_GAFL}
\end{figure}

\subsubsection{Inference Efficiency}
We report the time needed for sampling protein backbones of length 100 on an NVIDIA A100 GPU with published model weights and default settings as described in~\ref{sec:baselines}.
\label{sec:runtime}
\begin{table}[h]
\centering
\caption{Time needed for sampling protein backbones of length 100 on an NVIDIA A100 GPU.}
\label{tab:performance_comparison}
\begin{tabular}{@{}lcc@{}}
\toprule
Method       & Timesteps & Time per Structure (s) \\ \midrule
RFdiffusion       & 50         &  21.0               \\
FrameDiff    & 500         &   24.3             \\
FoldFlow     & 500       & 24.2                \\
FrameFlow   & 200 & 6.6 \\
\specialrule{1.2pt}{1.2pt}{1.2pt}
GAFL    & 200       & 8.8           \\
\bottomrule
\end{tabular}
\end{table}

\subsubsection{Training on PDB}
\label{sec:ckpt_hyperparams}
On the PDB dataset, we train both GAFL and FrameFlow for 5200 epochs, where one epoch is defined as one iteration over all 4776 clusters, which we define as in FrameDiff by 30\% sequence similarity.
The learning rate is increased in 50 warmup steps to 0.0002 and then kept constant for 3500 epochs.
From there we use a cosine-annealing schedule to decrease the learning rate to 0.0001 at epoch 5000.
From epoch $N_\text{train}=5000$ to $N_\text{train} + N_\text{select}=5200$ we employ our checkpointing criterion described in Section \ref{sec:training_details} evaluating secondary structures and storing checkpoints every second epoch.
We keep the $k=30$ best checkpoints and filter them for checkpoints with a secondary structure content deviation of less than $d_{\text{max}} < 0.2$.\\
In order to select the best checkpoint we evaluate all filtered checkpoints by sampling 40 backbones for each length in $\{100, 150, 200, 250, 300\}$ and choosing the checkpoint with the highest designability. We then run new and independent inference runs for all experiments we conduct on the selected checkpoint.

\subsubsection{Training on SCOPe (Ablation)}
\label{sec:scope_ablation}

On SCOPe, we train both GAFL and FrameFlow for 6500 epochs. We use a constant learning rate of 0.0001 for the whole procedure. From epoch $N_\text{train}=4000$ to $N_\text{train} + N_\text{select}=6500$ we use our ceckpointing criterion, storing checkpoints every 25 epochs. We keep the $k=10$ best checkpoints also filtering them for checkpoints with a secondary structure content deviation of less than $d_{\text{max}} < 0.2$. We evaluate each filtered checkpoint by sampling 10 backbones for each length in \{60,61,\dots,128\} and visualize the resulting designability distribution in Figure \ref{fig:scope_training}.

\begin{figure}[t]
\centering
\begin{subfigure}[t]{0.52\textwidth}
\centering
\includegraphics[width=\textwidth]{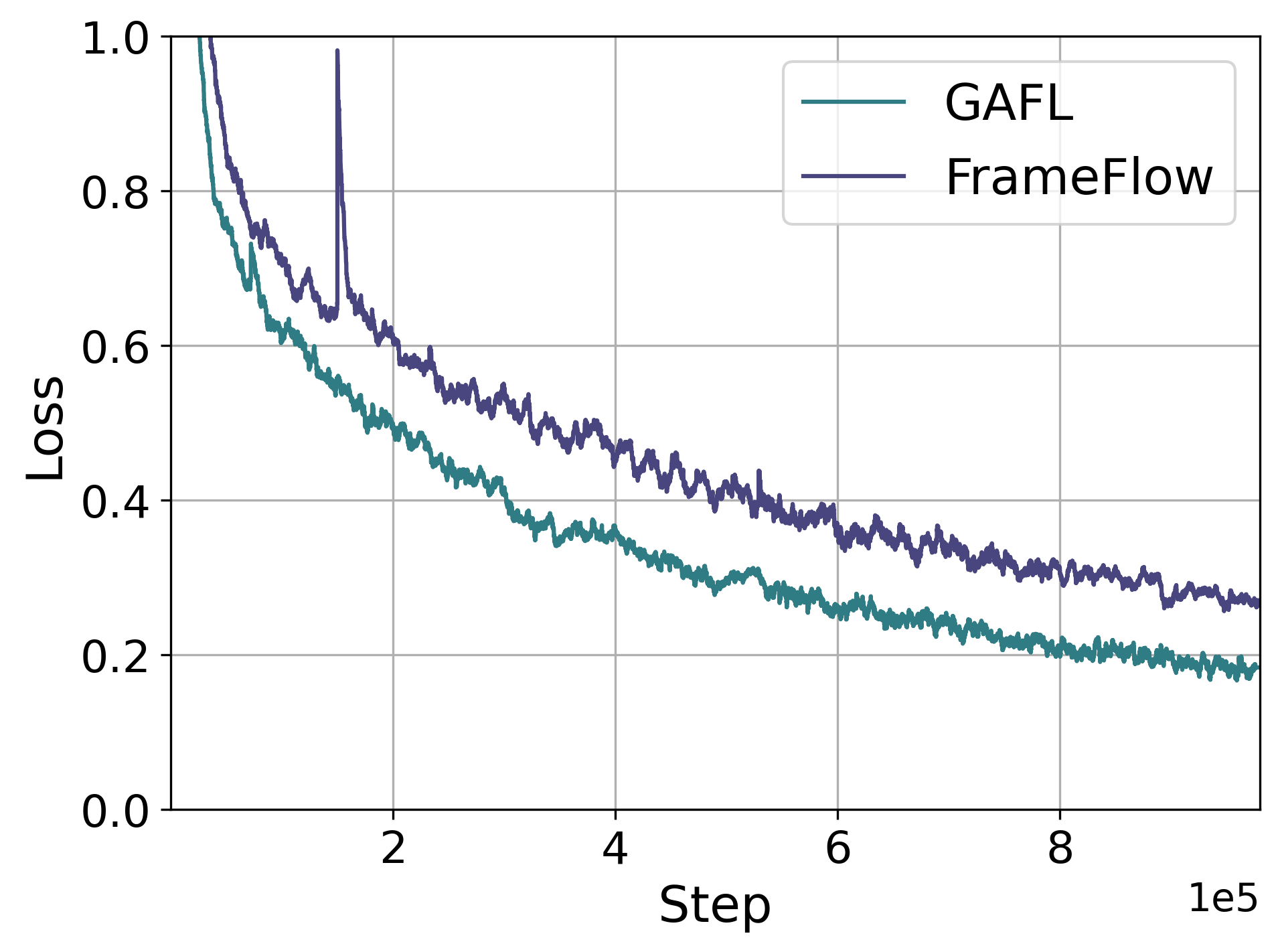}
\end{subfigure}
\hspace{0.04\textwidth}
\begin{subfigure}[t]{0.41\textwidth}
\centering
\includegraphics[width=\textwidth]{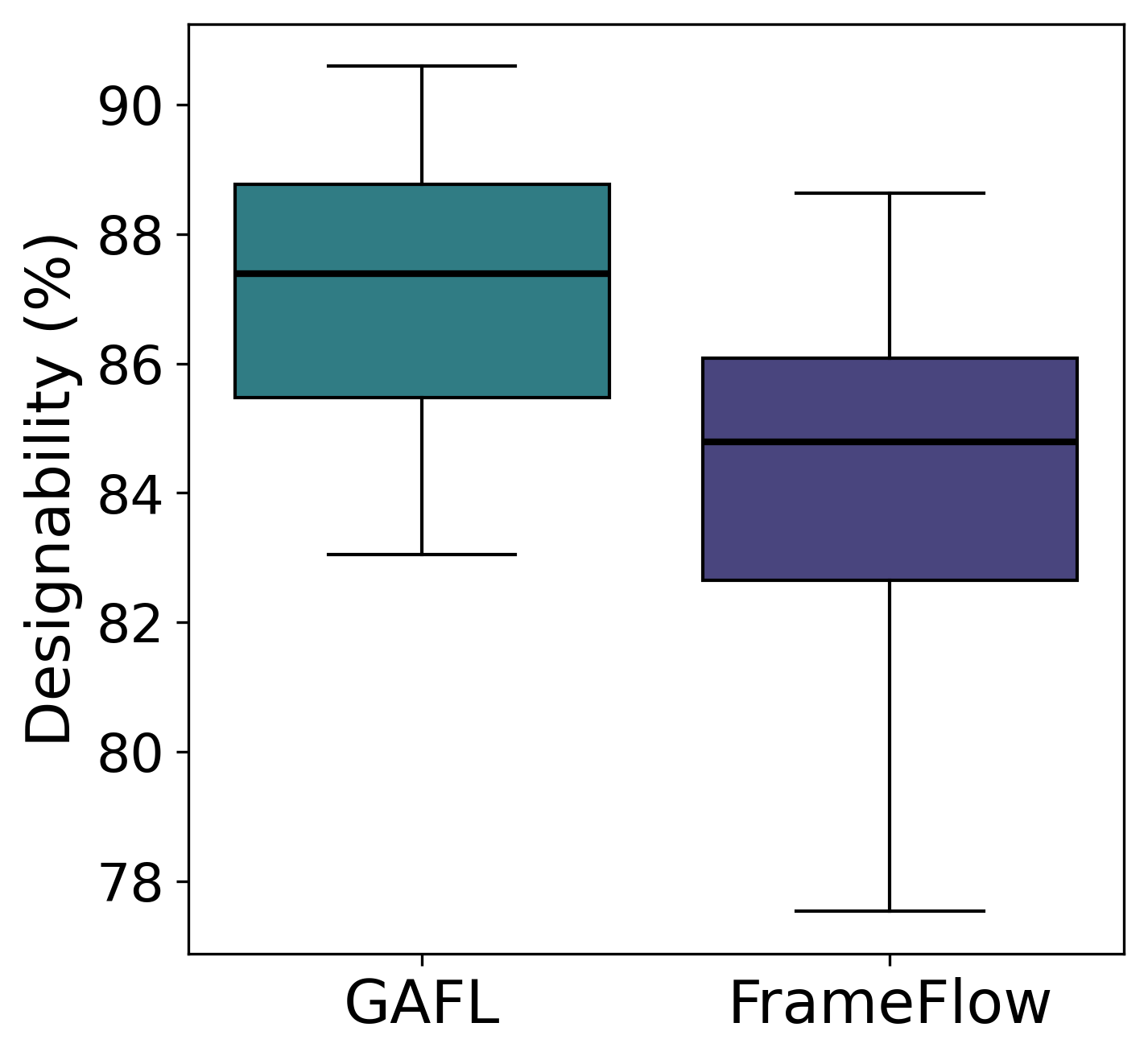} 
\end{subfigure}
\caption{Left: Total train loss averaged over flow matching times 0.5 to 0.75 on SCOPe for GAFL and FrameFlow. Right: Designabilities of 30 checkpoints sampled during three training runs on SCOPe for GAFL and retrained FrameFlow, respectively and filtered for a secondary structure content deviation of less than $d_{\text{max}} < 0.2$. For each checkpoint, we sample 10 backbones per length in \{60,61,\dots,128\}.}
\figlabel{scope_training}
\end{figure}

\subsubsection{Model hyperparameters}
\label{sec:model_hparam}

We report the most important hyperparameters of the CFA/IPA modules used in GAFL and FrameFlow respectively in Table \ref{table:model_hparam}. For an extensive list of hyperparameters we refer to the respective config files on GitHub. 

\begin{table}[h!]
    \centering
    \begin{tabular}{lcc}
    \toprule
    Parameter & GAFL & FrameFlow \\
    \midrule
    Node embedding size & 240 & 256\\
    Edge embedding size & 120 & 128\\
    Number of attention heads & 8 & 8\\
    Number of query/key points & 8 & 8\\
    Number of geometric value channels & 8 & 12\\
    Number of IPA/CFA blocks & 6 & 6\\
    \bottomrule
    \end{tabular}
    \caption{Comparison of the most important hyperparameters used for the CFA/IPA modules in GAFL and FrameFlow}
    \label{table:model_hparam}
\end{table}

The model hyperparameters of GAFL are choosen such that its total number of trainable parameters roughly equals that of FrameFlow as detailed in Table \ref{table:param_comp}.

\begin{table}[h!]
    \centering
    \begin{tabular}{lcc}
    \toprule
    Model component & GAFL & FrameFlow \\
    \midrule
    Embedding & 135 K & 150 K \\
    CFA/IPA & 9.2 M & 8.4 M \\
    Seq transformer & 4.5 M & 5.1 M \\
    Node update & 1.0 M & 1.2 M \\
    Edge update & 1.7 M & 1.9 M \\
    BB update & 169 K & 10 K \\
    \midrule
    Total & 16.7 M & 16.7 M \\
    \bottomrule
    \end{tabular}
    \caption{Comparison between the number of parameters of GAFL and FrameFlow, broken down into contributions from different components.}
    \label{table:param_comp}
\end{table}

\subsubsection{Memory consumption}

Training on SCOPe with the model hyperparameters listed in \ref{sec:model_hparam} and training hyperparameters given in \ref{sec:scope_ablation}, results in a GPU memory consumption of 48.2 GB for FrameFlow and 59.5 GB for GAFL. 

\subsection{Miscellaneous}
\subsubsection{Societal Impact}
\seclabel{societalimpact}
Societal impact is considered mostly positive, as \emph{de novo} protein design holds the promise to develop, for example, drugs against diseases, personalized therapies against cancer or also new nanomaterials, which out-weights potential risks, while of course security concerns remain, see e.g. \citet{baker-church-2024}.

\subsubsection{Implementation and Software Libraries}
\label{sec:implementation}

Our implementation is based on the implementation of FrameFlow~\citep{yim2023frameflow,yim2024improved}\footnote{The implementation of FrameFlow is published under the MIT license at \url{https://github.com/microsoft/protein-frame-flow}}. We will publish our code together with the camera ready version of this manuscript.
The implementation is in the Python~\citep{van1995python} programming language and uses the PyTorch framework~\citep{Paszke2019} and further dependencies of FrameFlow: Numpy~\citep{harris2020}, Hydra~\citep{Yadan2019Hydra}, and SciPy~\citep{2020SciPy-NMeth}.

%% file: 7_checklist.tex
\newpage
\section*{NeurIPS Paper Checklist}

\begin{enumerate}

\item {\bf Claims}
    \item[] Question: Do the main claims made in the abstract and introduction accurately reflect the paper's contributions and scope?
    \item[] Answer: \answerYes{} 
    \item[] Justification: In the abstract and introduction we made the following claims, which accurately reflect the papers contributions and scope:
    \begin{enumerate}
    \item Represent frames as elements of the projective geometric algebra
    \item Geometrically more expressive bilinear products of the geometric algebra
    \item Higher order message passing
    \item High designability and diversity of sampled protein backbones
    \item The proposed method closely captures the distribution of protein secondary structures, while other models with high designability often over-represent alpha helices
    \end{enumerate}

\item {\bf Limitations}
    \item[] Question: Does the paper discuss the limitations of the work performed by the authors?
    \item[] Answer: \answerYes{} 
    \item[] Justification: The limitations of the proposed method were discussed in a dedicated paragraph at the end of Section \secref{experiments}

\item {\bf Theory Assumptions and Proofs}
    \item[] Question: For each theoretical result, does the paper provide the full set of assumptions and a complete (and correct) proof?
    \item[] Answer: \answerNA{} 
    \item[] Justification: The paper does not introduce any novel theoretical results.

    \item {\bf Experimental Result Reproducibility}
    \item[] Question: Does the paper fully disclose all the information needed to reproduce the main experimental results of the paper to the extent that it affects the main claims and/or conclusions of the paper (regardless of whether the code and data are provided or not)?
    \item[] Answer: \answerYes{} 
    \item[] Justification: Full explanation of the experiments is provided in Section~\secref{experiments} and Appendix~\ref{sec:baselines}. The source code of the implementation will be published together with the camera ready version of the paper.

\item {\bf Open access to data and code}
    \item[] Question: Does the paper provide open access to the data and code, with sufficient instructions to faithfully reproduce the main experimental results, as described in supplemental material?
    \item[] Answer: \answerYes{} 
    \item[] Justification: The SCOPe dataset used for training the model is publicly available. The source code of the implementation of the method proposed in this work will be published together with the camera ready version of the paper.

\item {\bf Experimental Setting/Details}
    \item[] Question: Does the paper specify all the training and test details (e.g., data splits, hyperparameters, how they were chosen, type of optimizer, etc.) necessary to understand the results?
    \item[] Answer: \answerYes{} 
    \item[] Justification: We use the standard data set split of the SCOPe data set. For the baselines, we use the standard parameters defined in the respective publicly available official implementations.   The source code of the implementation of the method proposed in this work will be published together with the camera ready version of the paper.

\item {\bf Experiment Statistical Significance}
    \item[] Question: Does the paper report error bars suitably and correctly defined or other appropriate information about the statistical significance of the experiments?
    \item[] Answer: \answerYes{} 
    \item[] Justification: We report errors of the mean where applicable. Also we performed three additional complete training runs of the proposed method and the baseline FrameFlow with different seeds for verification. The results of this verification are provided in Table \ref{tab:ablation} and \ref{tab:gafl_ff_PDB}.

\item {\bf Experiments Compute Resources}
    \item[] Question: For each experiment, does the paper provide sufficient information on the computer resources (type of compute workers, memory, time of execution) needed to reproduce the experiments?
    \item[] Answer: \answerYes{} 
    \item[] Justification: We provided information on the hardware used for the experiments at the beginning of Section \secref{experiments}.
    
\item {\bf Code Of Ethics}
    \item[] Question: Does the research conducted in the paper conform, in every respect, with the NeurIPS Code of Ethics \url{https://neurips.cc/public/EthicsGuidelines}?
    \item[] Answer: \answerYes{} 
    \item[] Justification: No human subjects or data with privacy concerns was used. Training was only performed on publicly available standard academic datasets. Societal impact is considered mostly positive, as \emph{de novo} protein design holds the promise to develop drugs against diseases and personalized therapies against cancer, which out-weights potential risks, while of course security concerns remain, see e.g. Baker and Church(2024). The presented method does not enable the design of proteins with a certain function, so risk of misuse is considerably low.

\item {\bf Broader Impacts}
    \item[] Question: Does the paper discuss both potential positive societal impacts and negative societal impacts of the work performed?
    \item[] Answer: \answerYes{} 
    \item[] Justification: Societal impact is briefly discussed in Appendix \secref{societalimpact}, while we refer the reader to e.g Baker and Church (2024) for recent discussions of potential risks. The presented method does not enable the design of proteins with a certain function, so risk of misuse is considerably low.
    
\item {\bf Safeguards}
    \item[] Question: Does the paper describe safeguards that have been put in place for responsible release of data or models that have a high risk for misuse (e.g., pretrained language models, image generators, or scraped datasets)?
    \item[] Answer: \answerNA{} 
    \item[] Justification: The presented method does not enable the design of proteins with a certain function, so risk of misuse is considerably low.

\item {\bf Licenses for existing assets}
    \item[] Question: Are the creators or original owners of assets (e.g., code, data, models), used in the paper, properly credited and are the license and terms of use explicitly mentioned and properly respected?
    \item[] Answer: \answerYes{} 
    \item[] Justification: Our work is based on the implementation of FrameFlow, which was published under the MIT license which allows to create derivative works. The license terms will be followed when puiblishing our own implementations.

\item {\bf New Assets}
    \item[] Question: Are new assets introduced in the paper well documented and is the documentation provided alongside the assets?
    \item[] Answer: \answerNA{} 
    \item[] Justification: No new assets have been introduced.

\item {\bf Crowdsourcing and Research with Human Subjects}
    \item[] Question: For crowdsourcing experiments and research with human subjects, does the paper include the full text of instructions given to participants and screenshots, if applicable, as well as details about compensation (if any)? 
    \item[] Answer: \answerNA{} 
    \item[] Justification:No crowd sourcing or research with human subjects performed.

\item {\bf Institutional Review Board (IRB) Approvals or Equivalent for Research with Human Subjects}
    \item[] Question: Does the paper describe potential risks incurred by study participants, whether such risks were disclosed to the subjects, and whether Institutional Review Board (IRB) approvals (or an equivalent approval/review based on the requirements of your country or institution) were obtained?
    \item[] Answer: \answerNA{} 
    \item[] Justification: No insitutional review board approval required.

\end{enumerate}

\subsection*{References of NeurIPS Paper Checklist}
{
\small

David Baker and George Church.
\newblock Protein design meets biosecurity.
\newblock \emph{Science}, 383\penalty0 (6681):\penalty0 349--349, 2024.

}

%% file: paper_neurips_2024.bbl
\begin{thebibliography}{70}
\providecommand{\natexlab}[1]{#1}
\providecommand{\url}[1]{\texttt{#1}}
\expandafter\ifx\csname urlstyle\endcsname\relax
  \providecommand{\doi}[1]{doi: #1}\else
  \providecommand{\doi}{doi: \begingroup \urlstyle{rm}\Url}\fi

\bibitem[Anand and Achim(2022)]{anand2022protein}
Namrata Anand and Tudor Achim.
\newblock Protein structure and sequence generation with equivariant denoising
  diffusion probabilistic models.
\newblock \emph{arXiv preprint arXiv:2205.15019}, 2022.

\bibitem[Baek et~al.(2021)Baek, DiMaio, Anishchenko, Dauparas, Ovchinnikov,
  Lee, Wang, Cong, Kinch, Schaeffer, et~al.]{baek2021accurate}
Minkyung Baek, Frank DiMaio, Ivan Anishchenko, Justas Dauparas, Sergey
  Ovchinnikov, Gyu~Rie Lee, Jue Wang, Qian Cong, Lisa~N Kinch, R~Dustin
  Schaeffer, et~al.
\newblock Accurate prediction of protein structures and interactions using a
  three-track neural network.
\newblock \emph{Science}, 373\penalty0 (6557):\penalty0 871--876, 2021.

\bibitem[Baek et~al.(2023)Baek, Anishchenko, Humphreys, Cong, Baker, and
  DiMaio]{baek2023efficient}
Minkyung Baek, Ivan Anishchenko, Ian Humphreys, Qian Cong, David Baker, and
  Frank DiMaio.
\newblock Efficient and accurate prediction of protein structure using
  rosettafold2.
\newblock \emph{bioRxiv}, pages 2023--05, 2023.

\bibitem[Baker and Church(2024)]{baker-church-2024}
David Baker and George Church.
\newblock Protein design meets biosecurity.
\newblock \emph{Science}, 383\penalty0 (6681):\penalty0 349--349, 2024.

\bibitem[Baker et~al.(2024)Baker, Torres, Valle, Mackessy, Menzies, Casewell,
  Ahmadi, Burlet, Muratspahi{\'c}, Sappington, Overath, de~Torre, Ledergerber,
  Laustsen, Boddum, Bera, Kang, Brackenbrough, Cardoso, Crittenden, Edge,
  Decarreau, Ragotte, Pillai, Abedi, Han, Gerben, Murray, Skotheim, Stuart,
  Stewart, Fryer, and Jenkins]{torresnovo}
David Baker, Susana~V{\'a}zquez Torres, Melisa~Benard Valle, Stephen~P.
  Mackessy, Stefanie Menzies, Nicholas~R. Casewell, Shirin Ahmadi, Nick~J.
  Burlet, Edin Muratspahi{\'c}, Isaac Sappington, Max Overath, Esperanza~Rivera
  de~Torre, Jann Ledergerber, Andreas~Hougaard Laustsen, Kim Boddum, Asim~K.
  Bera, Alex Kang, Evans Brackenbrough, Iara Cardoso, Edouard Crittenden,
  Rebecca Edge, Justin Decarreau, Robert~J. Ragotte, Arvind Pillai, Mohamad~H.
  Abedi, Hannah Han, Stacey~R. Gerben, Analisa Murray, Rebecca Skotheim, Lynda
  Stuart, Lance Stewart, Thomas Fryer, and Timothy~Patrick Jenkins.
\newblock De novo designed proteins neutralize lethal snake venom toxins.
\newblock \emph{Research Square}, 2024.

\bibitem[Batatia et~al.(2022)Batatia, Kovacs, Simm, Ortner, and
  Cs{\'a}nyi]{batatia2022mace}
Ilyes Batatia, David~P Kovacs, Gregor Simm, Christoph Ortner, and G{\'a}bor
  Cs{\'a}nyi.
\newblock Mace: Higher order equivariant message passing neural networks for
  fast and accurate force fields.
\newblock \emph{Advances in Neural Information Processing Systems},
  35:\penalty0 11423--11436, 2022.

\bibitem[Batzner et~al.(2022)Batzner, Musaelian, Sun, Geiger, Mailoa,
  Kornbluth, Molinari, Smidt, and Kozinsky]{batzner2022}
Simon Batzner, Albert Musaelian, Lixin Sun, Mario Geiger, Jonathan~P Mailoa,
  Mordechai Kornbluth, Nicola Molinari, Tess~E Smidt, and Boris Kozinsky.
\newblock E (3)-equivariant graph neural networks for data-efficient and
  accurate interatomic potentials.
\newblock \emph{Nature communications}, 13\penalty0 (1):\penalty0 2453, 2022.

\bibitem[Bennett et~al.(2023)Bennett, Coventry, Goreshnik, Huang, Allen,
  Vafeados, Peng, Dauparas, Baek, Stewart, et~al.]{bennett2023improving}
Nathaniel~R Bennett, Brian Coventry, Inna Goreshnik, Buwei Huang, Aza Allen,
  Dionne Vafeados, Ying~Po Peng, Justas Dauparas, Minkyung Baek, Lance Stewart,
  et~al.
\newblock Improving de novo protein binder design with deep learning.
\newblock \emph{Nature Communications}, 14\penalty0 (1):\penalty0 2625, 2023.

\bibitem[Berman et~al.(2000)Berman, Westbrook, Feng, Gilliland, Bhat, Weissig,
  Shindyalov, and Bourne]{berman2000protein}
Helen~M Berman, John Westbrook, Zukang Feng, Gary Gilliland, Talapady~N Bhat,
  Helge Weissig, Ilya~N Shindyalov, and Philip~E Bourne.
\newblock The protein data bank.
\newblock \emph{Nucleic acids research}, 28\penalty0 (1):\penalty0 235--242,
  2000.

\bibitem[Bose et~al.(2024)Bose, Akhound-Sadegh, Huguet, FATRAS, Rector-Brooks,
  Liu, Nica, Korablyov, Bronstein, and Tong]{bose2023se}
Joey Bose, Tara Akhound-Sadegh, Guillaume Huguet, Kilian FATRAS, Jarrid
  Rector-Brooks, Cheng-Hao Liu, Andrei~Cristian Nica, Maksym Korablyov,
  Michael~M. Bronstein, and Alexander Tong.
\newblock {SE}(3)-stochastic flow matching for protein backbone generation.
\newblock In \emph{The Twelfth International Conference on Learning
  Representations}, 2024.

\bibitem[Brandstetter et~al.(2022)Brandstetter, Hesselink, van~der Pol,
  Bekkers, and Welling]{brandstetter2022geometric}
Johannes Brandstetter, Rob Hesselink, Elise van~der Pol, Erik~J Bekkers, and
  Max Welling.
\newblock Geometric and physical quantities improve e(3) equivariant message
  passing.
\newblock In \emph{International Conference on Learning Representations}, 2022.

\bibitem[Brandstetter et~al.(2023)Brandstetter, van~den Berg, Welling, and
  Gupta]{brandstetter2023clifford}
Johannes Brandstetter, Rianne van~den Berg, Max Welling, and Jayesh~K Gupta.
\newblock Clifford neural layers for {PDE} modeling.
\newblock In \emph{The Eleventh International Conference on Learning
  Representations}, 2023.

\bibitem[Brehmer et~al.(2023)Brehmer, de~Haan, Behrends, and
  Cohen]{brehmer2023geometric}
Johann Brehmer, Pim de~Haan, S{\"o}nke Behrends, and Taco Cohen.
\newblock Geometric algebra transformer.
\newblock In \emph{Advances in Neural Information Processing Systems},
  volume~37, 2023.

\bibitem[Bronstein et~al.(2017)Bronstein, Bruna, LeCun, Szlam, and
  Vandergheynst]{bronstein2017geometric}
Michael~M Bronstein, Joan Bruna, Yann LeCun, Arthur Szlam, and Pierre
  Vandergheynst.
\newblock Geometric deep learning: going beyond euclidean data.
\newblock \emph{IEEE Signal Processing Magazine}, 34\penalty0 (4):\penalty0
  18--42, 2017.

\bibitem[Bronstein et~al.(2021)Bronstein, Bruna, Cohen, and
  Veli{\v{c}}kovi{\'c}]{bronstein2021geometric}
Michael~M Bronstein, Joan Bruna, Taco Cohen, and Petar Veli{\v{c}}kovi{\'c}.
\newblock Geometric deep learning: Grids, groups, graphs, geodesics, and
  gauges.
\newblock \emph{arXiv preprint arXiv:2104.13478}, 2021.

\bibitem[Buchholz and Sommer(2008)]{buchholz2008clifford}
Sven Buchholz and Gerald Sommer.
\newblock On clifford neurons and clifford multi-layer perceptrons.
\newblock \emph{Neural Networks}, 21\penalty0 (7):\penalty0 925--935, 2008.

\bibitem[Cao et~al.(2020)Cao, Goreshnik, Coventry, Case, Miller, Kozodoy, Chen,
  Carter, Walls, Park, et~al.]{cao2020novo}
Longxing Cao, Inna Goreshnik, Brian Coventry, James~Brett Case, Lauren Miller,
  Lisa Kozodoy, Rita~E Chen, Lauren Carter, Alexandra~C Walls, Young-Jun Park,
  et~al.
\newblock De novo design of picomolar sars-cov-2 miniprotein inhibitors.
\newblock \emph{Science}, 370\penalty0 (6515):\penalty0 426--431, 2020.

\bibitem[Chandonia et~al.(2022)Chandonia, Guan, Lin, Yu, Fox, and
  Brenner]{chandonia2022scope}
John-Marc Chandonia, Lindsey Guan, Shiangyi Lin, Changhua Yu, Naomi~K Fox, and
  Steven~E Brenner.
\newblock Scope: improvements to the structural classification of
  proteins--extended database to facilitate variant interpretation and machine
  learning.
\newblock \emph{Nucleic acids research}, 50\penalty0 (D1):\penalty0 D553--D559,
  2022.

\bibitem[Chen and Lipman(2024)]{chen2023riemannian}
Ricky T.~Q. Chen and Yaron Lipman.
\newblock Flow matching on general geometries.
\newblock In \emph{The Twelfth International Conference on Learning
  Representations}, 2024.

\bibitem[Chen et~al.(2018)Chen, Rubanova, Bettencourt, and
  Duvenaud]{chen2018neural}
Ricky~TQ Chen, Yulia Rubanova, Jesse Bettencourt, and David~K Duvenaud.
\newblock Neural ordinary differential equations.
\newblock \emph{Advances in neural information processing systems}, 31, 2018.

\bibitem[Cohen(2021)]{cohen2021equivariant}
Taco Cohen.
\newblock \emph{Equivariant convolutional networks}.
\newblock PhD thesis, Taco Cohen, 2021.

\bibitem[Corso et~al.(2023)Corso, Jing, Barzilay, Jaakkola,
  et~al.]{corso2023diffdock}
Gabriele Corso, Bowen Jing, Regina Barzilay, Tommi Jaakkola, et~al.
\newblock Diffdock: Diffusion steps, twists, and turns for molecular docking.
\newblock In \emph{International Conference on Learning Representations (ICLR
  2023)}, 2023.

\bibitem[Dauparas et~al.(2022)Dauparas, Anishchenko, Bennett, Bai, Ragotte,
  Milles, Wicky, Courbet, de~Haas, Bethel, Leung, Huddy, Pellock, Tischer,
  Chan, Koepnick, Nguyen, Kang, Sankaran, Bera, King, and
  Baker]{dauparas2022protmpnn}
J.~Dauparas, I.~Anishchenko, N.~Bennett, H.~Bai, R.~J. Ragotte, L.~F. Milles,
  B.~I.~M. Wicky, A.~Courbet, R.~J. de~Haas, N.~Bethel, P.~J.~Y. Leung, T.~F.
  Huddy, S.~Pellock, D.~Tischer, F.~Chan, B.~Koepnick, H.~Nguyen, A.~Kang,
  B.~Sankaran, A.~K. Bera, N.~P. King, and D.~Baker.
\newblock Robust deep learning-based protein sequence design using
  {ProteinMPNN}.
\newblock \emph{Science}, 378\penalty0 (6615):\penalty0 49--56, 2022.

\bibitem[Doran and Lasenby(2003)]{doran2003geometric}
Chris Doran and Anthony Lasenby.
\newblock \emph{Geometric algebra for physicists}.
\newblock Cambridge University Press, 2003.

\bibitem[Dorst and De~Keninck(2020)]{dorst2020guided}
Leo Dorst and Steven De~Keninck.
\newblock A guided tour to the plane-based geometric algebra pga.
\newblock 2020.
\newblock URL \url{https://bivector.net/PGA4CS.html}.

\bibitem[Dorst et~al.(2007)Dorst, Fontijne, and Mann]{dorst2007geometric}
Leo Dorst, Daniel Fontijne, and Stephen Mann.
\newblock \emph{Geometric Algebra for Computer Science: An Object-Oriented
  Approach to Geometry}.
\newblock Morgan Kaufmann Publishers Inc., San Francisco, CA, USA, 1st edition,
  2007.
\newblock ISBN 0123694655.

\bibitem[Fox et~al.(2014)Fox, Brenner, and Chandonia]{fox2014scope}
Naomi~K Fox, Steven~E Brenner, and John-Marc Chandonia.
\newblock Scope: Structural classification of proteins—extended, integrating
  scop and astral data and classification of new structures.
\newblock \emph{Nucleic acids research}, 42\penalty0 (D1):\penalty0 D304--D309,
  2014.

\bibitem[Gasteiger et~al.(2020)Gasteiger, Groß, and
  Günnemann]{gasteiger2019directional}
Johannes Gasteiger, Janek Groß, and Stephan Günnemann.
\newblock Directional message passing for molecular graphs.
\newblock In \emph{International Conference on Learning Representations}, 2020.

\bibitem[Gunn(2020)]{gunn2020projective}
Charles~G. Gunn.
\newblock Projective geometric algebra: A new framework for doing euclidean
  geometry.
\newblock \emph{arXiv preprint arXiv:1901.05873}, 2020.

\bibitem[Harris et~al.(2020)Harris, Millman, van~der Walt, Gommers, Virtanen,
  Cournapeau, Wieser, Taylor, Berg, Smith, Kern, Picus, Hoyer, van Kerkwijk,
  Brett, Haldane, del R{\'{i}}o, Wiebe, Peterson, G{\'{e}}rard-Marchant,
  Sheppard, Reddy, Weckesser, Abbasi, Gohlke, and Oliphant]{harris2020}
Charles~R. Harris, K.~Jarrod Millman, St{\'{e}}fan~J. van~der Walt, Ralf
  Gommers, Pauli Virtanen, David Cournapeau, Eric Wieser, Julian Taylor,
  Sebastian Berg, Nathaniel~J. Smith, Robert Kern, Matti Picus, Stephan Hoyer,
  Marten~H. van Kerkwijk, Matthew Brett, Allan Haldane, Jaime~Fern{\'{a}}ndez
  del R{\'{i}}o, Mark Wiebe, Pearu Peterson, Pierre G{\'{e}}rard-Marchant,
  Kevin Sheppard, Tyler Reddy, Warren Weckesser, Hameer Abbasi, Christoph
  Gohlke, and Travis~E. Oliphant.
\newblock Array programming with {NumPy}.
\newblock \emph{Nature}, 585\penalty0 (7825):\penalty0 357--362, September
  2020.

\bibitem[Hestenes and Sobczyk(2012)]{hestenes2012clifford}
David Hestenes and Garret Sobczyk.
\newblock \emph{Clifford algebra to geometric calculus: a unified language for
  mathematics and physics}, volume~5.
\newblock Springer Science \& Business Media, 2012.

\bibitem[Jumper et~al.(2021)Jumper, Evans, Pritzel, Green, Figurnov,
  Ronneberger, Tunyasuvunakool, Bates, {\v Z}{\'\i}dek, Potapenko, Bridgland,
  Meyer, Kohl, Ballard, Cowie, Romera-Paredes, Nikolov, Jain, Adler, Back,
  Petersen, Reiman, Clancy, Zielinski, Steinegger, Pacholska, Berghammer,
  Bodenstein, Silver, Vinyals, Senior, Kavukcuoglu, Kohli, and
  Hassabis]{jumper2021highly}
John Jumper, Richard Evans, Alexander Pritzel, Tim Green, Michael Figurnov,
  Olaf Ronneberger, Kathryn Tunyasuvunakool, Russ Bates, Augustin {\v
  Z}{\'\i}dek, Anna Potapenko, Alex Bridgland, Clemens Meyer, Simon A~A Kohl,
  Andrew~J Ballard, Andrew Cowie, Bernardino Romera-Paredes, Stanislav Nikolov,
  Rishub Jain, Jonas Adler, Trevor Back, Stig Petersen, David Reiman, Ellen
  Clancy, Michal Zielinski, Martin Steinegger, Michalina Pacholska, Tamas
  Berghammer, Sebastian Bodenstein, David Silver, Oriol Vinyals, Andrew~W
  Senior, Koray Kavukcuoglu, Pushmeet Kohli, and Demis Hassabis.
\newblock Highly accurate protein structure prediction with {AlphaFold}.
\newblock \emph{Nature}, 596\penalty0 (7873):\penalty0 583--589, August 2021.

\bibitem[Kabsch and Sander(1983)]{kabsch1983dictionary}
Wolfgang Kabsch and Christian Sander.
\newblock Dictionary of protein secondary structure: pattern recognition of
  hydrogen-bonded and geometrical features.
\newblock \emph{Biopolymers: Original Research on Biomolecules}, 22\penalty0
  (12):\penalty0 2577--2637, 1983.

\bibitem[Klein et~al.(2023)Klein, Kr\"{a}mer, and Noe]{klein2023equivariant}
Leon Klein, Andreas Kr\"{a}mer, and Frank Noe.
\newblock Equivariant flow matching.
\newblock In A.~Oh, T.~Naumann, A.~Globerson, K.~Saenko, M.~Hardt, and
  S.~Levine, editors, \emph{Advances in Neural Information Processing Systems},
  volume~36, pages 59886--59910. Curran Associates, Inc., 2023.

\bibitem[Leaver-Fay et~al.(2011)Leaver-Fay, Tyka, Lewis, Lange, Thompson,
  Jacak, Kaufman, Renfrew, Smith, Sheffler, et~al.]{leaver2011rosetta3}
Andrew Leaver-Fay, Michael Tyka, Steven~M Lewis, Oliver~F Lange, James
  Thompson, Ron Jacak, Kristian~W Kaufman, P~Douglas Renfrew, Colin~A Smith,
  Will Sheffler, et~al.
\newblock Rosetta3: an object-oriented software suite for the simulation and
  design of macromolecules.
\newblock In \emph{Methods in enzymology}, volume 487, pages 545--574.
  Elsevier, 2011.

\bibitem[Liao and Smidt(2023)]{liao2023equiformer}
Yi-Lun Liao and Tess Smidt.
\newblock Equiformer: Equivariant graph attention transformer for 3d atomistic
  graphs.
\newblock In \emph{The Eleventh International Conference on Learning
  Representations}, 2023.

\bibitem[Lin and AlQuraishi(2023)]{lin2023genie}
Yeqing Lin and Mohammed AlQuraishi.
\newblock Generating novel, designable, and diverse protein structures by
  equivariantly diffusing oriented residue clouds.
\newblock \emph{arXiv preprint arXiv:2301.12485}, 2023.

\bibitem[Lin et~al.(2023)Lin, Akin, Rao, Hie, Zhu, Lu, Smetanin, Verkuil,
  Kabeli, Shmueli, et~al.]{lin2023evolutionary}
Zeming Lin, Halil Akin, Roshan Rao, Brian Hie, Zhongkai Zhu, Wenting Lu, Nikita
  Smetanin, Robert Verkuil, Ori Kabeli, Yaniv Shmueli, et~al.
\newblock Evolutionary-scale prediction of atomic-level protein structure with
  a language model.
\newblock \emph{Science}, 379\penalty0 (6637):\penalty0 1123--1130, 2023.

\bibitem[Lipman et~al.(2023)Lipman, Chen, Ben-Hamu, Nickel, and
  Le]{lipman2022flow}
Yaron Lipman, Ricky~TQ Chen, Heli Ben-Hamu, Maximilian Nickel, and Matt Le.
\newblock Flow matching for generative modeling.
\newblock \emph{International Conference on Learning Representations}, 2023.

\bibitem[Lounesto(2001)]{lounesto2001clifford}
Pertti Lounesto.
\newblock Clifford algebras and spinors.
\newblock In \emph{Clifford algebras and their applications in mathematical
  physics}, pages 25--37. Springer, 2001.

\bibitem[Lundholm and Svensson(2009)]{lundholm2009clifford}
Douglas Lundholm and Lars Svensson.
\newblock Clifford algebra, geometric algebra, and applications.
\newblock \emph{arXiv preprint arXiv:0907.5356}, 2009.

\bibitem[Mao et~al.(2023)Mao, Zhu, Chen, and Shen]{mao2023modeling}
Weian Mao, Muzhi Zhu, Hao Chen, and Chunhua Shen.
\newblock Modeling protein structure using geometric vector field networks.
\newblock \emph{bioRxiv}, pages 2023--05, 2023.

\bibitem[Olshefsky et~al.(2022)Olshefsky, Richardson, Pun, and
  King]{olshefsky2022engineering}
Audrey Olshefsky, Christian Richardson, Suzie~H Pun, and Neil~P King.
\newblock Engineering self-assembling protein nanoparticles for therapeutic
  delivery.
\newblock \emph{Bioconjugate chemistry}, 33\penalty0 (11):\penalty0 2018--2034,
  2022.

\bibitem[Ovchinnikov and Huang(2021)]{ovchinnikov2021structure}
Sergey Ovchinnikov and Po-Ssu Huang.
\newblock Structure-based protein design with deep learning.
\newblock \emph{Current opinion in chemical biology}, 65:\penalty0 136--144,
  2021.

\bibitem[Paszke et~al.(2019)Paszke, Gross, Massa, Lerer, Bradbury, Chanan,
  Killeen, Lin, Gimelshein, Antiga, Desmaison, Kopf, Yang, DeVito, Raison,
  Tejani, Chilamkurthy, Steiner, Fang, Bai, and Chintala]{Paszke2019}
Adam Paszke, Sam Gross, Francisco Massa, Adam Lerer, James Bradbury, Gregory
  Chanan, Trevor Killeen, Zeming Lin, Natalia Gimelshein, Luca Antiga, Alban
  Desmaison, Andreas Kopf, Edward Yang, Zachary DeVito, Martin Raison, Alykhan
  Tejani, Sasank Chilamkurthy, Benoit Steiner, Lu~Fang, Junjie Bai, and Soumith
  Chintala.
\newblock Pytorch: An imperative style, high-performance deep learning library.
\newblock In \emph{Advances in Neural Information Processing Systems}. 2019.

\bibitem[Pearson and Bisset(1994)]{pearson1994neural}
J.K. Pearson and D.L. Bisset.
\newblock Neural networks in the clifford domain.
\newblock In \emph{Proceedings of 1994 IEEE International Conference on Neural
  Networks (ICNN'94)}, volume~3, pages 1465--1469 vol.3, 1994.

\bibitem[Roney and Ovchinnikov(2022)]{roney2022state}
James~P Roney and Sergey Ovchinnikov.
\newblock State-of-the-art estimation of protein model accuracy using
  alphafold.
\newblock \emph{Physical Review Letters}, 129\penalty0 (23):\penalty0 238101,
  2022.

\bibitem[Ruhe et~al.(2023{\natexlab{a}})Ruhe, Brandstetter, and
  Forr{\'e}]{ruhe2023clifford}
David Ruhe, Johannes Brandstetter, and Patrick Forr{\'e}.
\newblock Clifford group equivariant neural networks.
\newblock In \emph{Thirty-seventh Conference on Neural Information Processing
  Systems}, 2023{\natexlab{a}}.

\bibitem[Ruhe et~al.(2023{\natexlab{b}})Ruhe, Gupta, De~Keninck, Welling, and
  Brandstetter]{ruhe2023geometric}
David Ruhe, Jayesh~K Gupta, Steven De~Keninck, Max Welling, and Johannes
  Brandstetter.
\newblock Geometric clifford algebra networks.
\newblock In \emph{International Conference on Machine Learning}, pages
  29306--29337. PMLR, 2023{\natexlab{b}}.

\bibitem[Satorras et~al.(2021)Satorras, Hoogeboom, and
  Welling]{satorras2021gnn}
V\'{\i}ctor~Garcia Satorras, Emiel Hoogeboom, and Max Welling.
\newblock E(n) equivariant graph neural networks.
\newblock In Marina Meila and Tong Zhang, editors, \emph{Proceedings of the
  38th International Conference on Machine Learning}, volume 139 of
  \emph{Proceedings of Machine Learning Research}, pages 9323--9332. PMLR,
  18--24 Jul 2021.

\bibitem[Schymkowitz et~al.(2005)Schymkowitz, Borg, Stricher, Nys, Rousseau,
  and Serrano]{schymkowitz2005foldx}
Joost Schymkowitz, Jesper Borg, Francois Stricher, Robby Nys, Frederic
  Rousseau, and Luis Serrano.
\newblock The foldx web server: an online force field.
\newblock \emph{Nucleic acids research}, 33\penalty0 (suppl\_2):\penalty0
  W382--W388, 2005.

\bibitem[Simeon and De~Fabritiis(2023)]{simeon2023tensornet}
Guillem Simeon and Gianni De~Fabritiis.
\newblock Tensornet: Cartesian tensor representations for efficient learning of
  molecular potentials.
\newblock In A.~Oh, T.~Naumann, A.~Globerson, K.~Saenko, M.~Hardt, and
  S.~Levine, editors, \emph{Advances in Neural Information Processing Systems},
  volume~36, pages 37334--37353. Curran Associates, Inc., 2023.

\bibitem[Sohl-Dickstein et~al.(2015)Sohl-Dickstein, Weiss, Maheswaranathan, and
  Ganguli]{sohl2015deep}
Jascha Sohl-Dickstein, Eric Weiss, Niru Maheswaranathan, and Surya Ganguli.
\newblock Deep unsupervised learning using nonequilibrium thermodynamics.
\newblock In \emph{International conference on machine learning}, pages
  2256--2265. PMLR, 2015.

\bibitem[Song et~al.(2020)Song, Sohl-Dickstein, Kingma, Kumar, Ermon, and
  Poole]{song2020score}
Yang Song, Jascha Sohl-Dickstein, Diederik~P Kingma, Abhishek Kumar, Stefano
  Ermon, and Ben Poole.
\newblock Score-based generative modeling through stochastic differential
  equations.
\newblock \emph{arXiv preprint arXiv:2011.13456}, 2020.

\bibitem[Thomas et~al.(2018)Thomas, Smidt, Kearnes, Yang, Li, Kohlhoff, and
  Riley]{thomas2018tensor}
Nathaniel Thomas, Tess Smidt, Steven Kearnes, Lusann Yang, Li~Li, Kai Kohlhoff,
  and Patrick Riley.
\newblock Tensor field networks: Rotation- and translation-equivariant neural
  networks for 3d point clouds.
\newblock \emph{arXiv preprint arXiv:1802.08219}, 2018.

\bibitem[Tong et~al.(2024)Tong, Fatras, Malkin, Huguet, Zhang, Rector-Brooks,
  Wolf, and Bengio]{tong2024improving}
Alexander Tong, Kilian Fatras, Nikolay Malkin, Guillaume Huguet, Yanlei Zhang,
  Jarrid Rector-Brooks, Guy Wolf, and Yoshua Bengio.
\newblock Improving and generalizing flow-based generative models with
  minibatch optimal transport.
\newblock \emph{arXiv preprint arXiv:2302.00482}, 2024.

\bibitem[Trippe et~al.(2022)Trippe, Yim, Tischer, Baker, Broderick, Barzilay,
  and Jaakkola]{trippe2022diffusion}
Brian~L Trippe, Jason Yim, Doug Tischer, David Baker, Tamara Broderick, Regina
  Barzilay, and Tommi Jaakkola.
\newblock Diffusion probabilistic modeling of protein backbones in 3d for the
  motif-scaffolding problem.
\newblock \emph{arXiv preprint arXiv:2206.04119}, 2022.

\bibitem[Van~Kempen et~al.(2024)Van~Kempen, Kim, Tumescheit, Mirdita, Lee,
  Gilchrist, S{\"o}ding, and Steinegger]{van2024fast}
Michel Van~Kempen, Stephanie~S Kim, Charlotte Tumescheit, Milot Mirdita,
  Jeongjae Lee, Cameron~LM Gilchrist, Johannes S{\"o}ding, and Martin
  Steinegger.
\newblock Fast and accurate protein structure search with foldseek.
\newblock \emph{Nature Biotechnology}, 42\penalty0 (2):\penalty0 243--246,
  2024.

\bibitem[Van~Rossum and Drake~Jr(1995)]{van1995python}
Guido Van~Rossum and Fred~L Drake~Jr.
\newblock \emph{Python reference manual}.
\newblock Centrum voor Wiskunde en Informatica Amsterdam, 1995.

\bibitem[Vaswani et~al.(2017)Vaswani, Shazeer, Parmar, Uszkoreit, Jones, Gomez,
  Kaiser, and Polosukhin]{vaswani2017attention}
Ashish Vaswani, Noam Shazeer, Niki Parmar, Jakob Uszkoreit, Llion Jones,
  Aidan~N Gomez, {\L}ukasz Kaiser, and Illia Polosukhin.
\newblock Attention is all you need.
\newblock \emph{Advances in neural information processing systems}, 30, 2017.

\bibitem[Virtanen et~al.(2020)Virtanen, Gommers, Oliphant, Haberland, Reddy,
  Cournapeau, Burovski, Peterson, Weckesser, Bright, {van der Walt}, Brett,
  Wilson, Millman, Mayorov, Nelson, Jones, Kern, Larson, Carey, Polat, Feng,
  Moore, {VanderPlas}, Laxalde, Perktold, Cimrman, Henriksen, Quintero, Harris,
  Archibald, Ribeiro, Pedregosa, {van Mulbregt}, and {SciPy 1.0
  Contributors}]{2020SciPy-NMeth}
Pauli Virtanen, Ralf Gommers, Travis~E. Oliphant, Matt Haberland, Tyler Reddy,
  David Cournapeau, Evgeni Burovski, Pearu Peterson, Warren Weckesser, Jonathan
  Bright, St{\'e}fan~J. {van der Walt}, Matthew Brett, Joshua Wilson, K.~Jarrod
  Millman, Nikolay Mayorov, Andrew R.~J. Nelson, Eric Jones, Robert Kern, Eric
  Larson, C~J Carey, {\.I}lhan Polat, Yu~Feng, Eric~W. Moore, Jake
  {VanderPlas}, Denis Laxalde, Josef Perktold, Robert Cimrman, Ian Henriksen,
  E.~A. Quintero, Charles~R. Harris, Anne~M. Archibald, Ant{\^o}nio~H. Ribeiro,
  Fabian Pedregosa, Paul {van Mulbregt}, and {SciPy 1.0 Contributors}.
\newblock {{SciPy} 1.0: Fundamental Algorithms for Scientific Computing in
  Python}.
\newblock \emph{Nature Methods}, 17:\penalty0 261--272, 2020.

\bibitem[Wang and Zhang(2022)]{wang2022graph}
Xiyuan Wang and Muhan Zhang.
\newblock Graph neural network with local frame for molecular potential energy
  surface.
\newblock In \emph{Learning on Graphs Conference}, pages 19--1. PMLR, 2022.

\bibitem[Watson et~al.(2023)Watson, Juergens, Bennett, Trippe, Yim, Eisenach,
  Ahern, Borst, Ragotte, Milles, et~al.]{watson2023novo}
Joseph~L Watson, David Juergens, Nathaniel~R Bennett, Brian~L Trippe, Jason
  Yim, Helen~E Eisenach, Woody Ahern, Andrew~J Borst, Robert~J Ragotte, Lukas~F
  Milles, et~al.
\newblock De novo design of protein structure and function with rfdiffusion.
\newblock \emph{Nature}, pages 1--3, 2023.

\bibitem[Webb and Sali(2016)]{webb2016comparative}
Benjamin Webb and Andrej Sali.
\newblock Comparative protein structure modeling using modeller.
\newblock \emph{Current protocols in bioinformatics}, 54\penalty0 (1):\penalty0
  5--6, 2016.

\bibitem[Wu et~al.(2024)Wu, Yang, van~den Berg, Alamdari, Zou, Lu, and
  Amini]{wu2024protein}
Kevin~E Wu, Kevin~K Yang, Rianne van~den Berg, Sarah Alamdari, James~Y Zou,
  Alex~X Lu, and Ava~P Amini.
\newblock Protein structure generation via folding diffusion.
\newblock \emph{Nature Communications}, 15\penalty0 (1):\penalty0 1059, 2024.

\bibitem[Yadan(2019)]{Yadan2019Hydra}
Omry Yadan.
\newblock Hydra - a framework for elegantly configuring complex applications.
\newblock Github, 2019.
\newblock URL \url{https://github.com/facebookresearch/hydra}.

\bibitem[Yim et~al.(2023{\natexlab{a}})Yim, Campbell, Foong, Gastegger,
  Jim{\'e}nez-Luna, Lewis, Satorras, Veeling, Barzilay, Jaakkola,
  et~al.]{yim2023frameflow}
Jason Yim, Andrew Campbell, Andrew~YK Foong, Michael Gastegger, Jos{\'e}
  Jim{\'e}nez-Luna, Sarah Lewis, Victor~Garcia Satorras, Bastiaan~S Veeling,
  Regina Barzilay, Tommi Jaakkola, et~al.
\newblock Fast protein backbone generation with se (3) flow matching.
\newblock \emph{arXiv preprint arXiv:2310.05297}, 2023{\natexlab{a}}.

\bibitem[Yim et~al.(2023{\natexlab{b}})Yim, Trippe, De~Bortoli, Mathieu,
  Doucet, Barzilay, and Jaakkola]{yim2023framediff}
Jason Yim, Brian~L Trippe, Valentin De~Bortoli, Emile Mathieu, Arnaud Doucet,
  Regina Barzilay, and Tommi Jaakkola.
\newblock Se (3) diffusion model with application to protein backbone
  generation.
\newblock \emph{arXiv preprint arXiv:2302.02277}, 2023{\natexlab{b}}.

\bibitem[Yim et~al.(2024)Yim, Campbell, Mathieu, Foong, Gastegger,
  Jimenez-Luna, Lewis, Satorras, Veeling, Noe, Barzilay, and
  Jaakkola]{yim2024improved}
Jason Yim, Andrew Campbell, Emile Mathieu, Andrew Y.~K. Foong, Michael
  Gastegger, Jose Jimenez-Luna, Sarah Lewis, Victor~Garcia Satorras,
  Bastiaan~S. Veeling, Frank Noe, Regina Barzilay, and Tommi Jaakkola.
\newblock Improved motif-scaffolding with {SE}(3) flow matching.
\newblock \emph{Transactions on Machine Learning Research}, 2024.
\newblock ISSN 2835-8856.

\bibitem[Zhang and Skolnick(2004)]{zhang2004scoring}
Yang Zhang and Jeffrey Skolnick.
\newblock Scoring function for automated assessment of protein structure
  template quality.
\newblock \emph{Proteins: Structure, Function, and Bioinformatics}, 57\penalty0
  (4):\penalty0 702--710, 2004.

\end{thebibliography}
